\documentclass[fleqn,10pt]{style/wlscirep}

\usepackage{hyperref}
\usepackage{url}
\usepackage{graphicx}
\usepackage{multirow}
\usepackage{booktabs} 
\usepackage[utf8]{inputenc}
\usepackage{tcolorbox}
\usepackage{xcolor} 
\usepackage[T1]{fontenc}    
\usepackage{booktabs}       
\usepackage{amsfonts}       
\usepackage{nicefrac}       
\usepackage{microtype}      
\usepackage[table]{xcolor} 
\usepackage{tikz}          
\usepackage{collcell}      
\usepackage{booktabs} 
\usepackage{float}
\usepackage{array}

\title{Pathology-CoT: Learning Visual Chain-of-Thought Agent from Expert Whole Slide Image Diagnosis Behavior}

\author[1,*]{Sheng Wang}
\author[2,*]{Ruiming Wu}
\author[5]{Charles Herndon}
\author[3]{Yihang Liu}
\author[1]{Shunsuke Koga}
\author[4]{Jeanne Shen}
\author[1,2,$\dagger$]{Zhi Huang}
\affil[1]{Department of Pathology and Laboratory Medicine, University of Pennsylvania}
\affil[2]{Department of Biostatistics,
Epidemiology \& Informatics, University of Pennsylvania}
\affil[3]{Department of Electrical and System Engineering, University of Pennsylvania}
\affil[4]{Department of Pathology, Stanford University}
\affil[5]{Department of Pathology, University of California at San Francisco}

\affil[*]{These authors contributed equally to this work}

\affil[$\dagger$]{To whom the correspondence should be addressed: Zhi Huang (\href{zhi.huang@pennmedicine.upenn.edu}{zhi.huang@pennmedicine.upenn.edu})}

\begin{abstract}

Diagnosing a whole-slide image is an interactive, multi-stage process of changing magnification and moving between fields. 
Although recent pathology foundation models demonstrated superior performances, practical agentic systems that decide what field to examine next, adjust magnification, and deliver explainable diagnoses are still lacking.
Such limitation is largely bottlenecked by data -- scalable, clinically aligned supervision of expert viewing behavior that is tacit and experience‑based, not documented in textbooks or internet, and therefore absent from LLM training.
Here we introduce \textbf{Pathology-CoT}, a framework designed to address this challenge through three key breakthroughs. First, the \textbf{AI Session Recorder} seamlessly integrates with standard whole-slide image (WSI) viewers to unobtrusively record routine navigation and convert the viewer logs into standardized behavioral commands (inspect/peek at discrete magnifications) and bounding boxes. Second,
a lightweight human-in-the-loop review turns AI-drafted rationales for behavioral commands into the \textbf{Pathology-CoT dataset}, a form of paired ``where to look” and ``why it matters”, enabling six-fold faster labeling compared to manual constructing such Chain-of-Thought dataset.
Using this behavioral data, we build \textbf{Pathology-o3}, a two-stage agent that first proposes important regions of interest and then performs behavior-guided reasoning. 
On the gastrointestinal lymph-node metastasis detection task, our method achieved 100.0\% recall on the internal validation dataset from Stanford Medicine and 97.6\% recall on an independent external validation dataset from Sweden, exceeding the state-of-the-art OpenAI o3 model and generalizing across backbones. 
To our knowledge, Pathology-CoT constitutes one of the first behavior-grounded agentic systems in pathology.
Turning everyday viewer logs into scalable, expert‑validated supervision, our framework makes agentic pathology practical and establishes a path to human‑aligned, upgradeable clinical AI.
\end{abstract}

\begin{document}
\flushbottom
\maketitle

\label{sec:introduction}

\section{Introduction}

The advancement of digital pathology and its integration of artificial intelligence (AI) have become deeply intertwined over the past two decades \cite{Gurcan2009HistopathReview,Echle2021DeepLearningPathology}.
This progress enabled the first generation of computer-aided diagnosis tools, some of which are now incorporated into clinical workflows for tasks such as tumor detection and cell counting \cite{Campanella2019ClinicalGradeCompPath,EhteshamiBejnordi2017JAMA,Coudray2018NSCLCHistopathology,Veta2015MitosisDetectionReview}.
These early successes demonstrated the value of AI in some of well-defined problems within pathology.
More recently, the field has been transformed by the emergence of powerful self-supervised pathology foundation models \cite{Xu2024WholeSlideFoundation,Vorontsov2023Virchow,Chen2024towards} alongside large language models (LLMs) and vision language models (VLMs). 
Together, these models combine visual understanding with language-level reasoning, opening the way to the next frontier: AI pathology agents.

\begin{figure}[hbtp]
    \centering
    \includegraphics[width=0.83\linewidth]{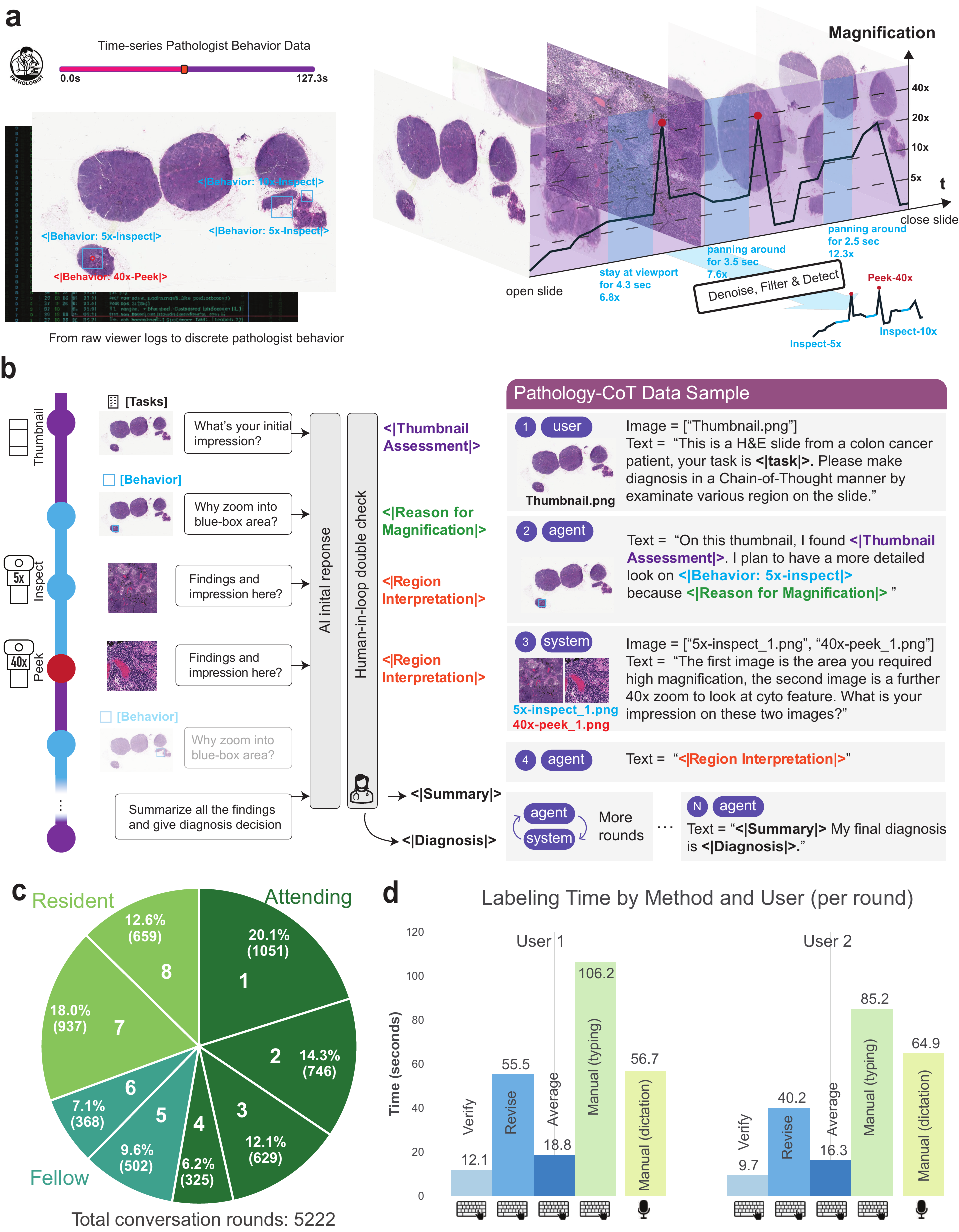} 
    \caption{The \textbf{AI Session Recorder} framework for converting expert viewing behavior into agent-ready training data.
    \textbf{a} The AI Session Recorder addresses the core data challenge: raw, continuous user logs are too noisy and complex for AI to learn from. The recorder's algorithm transforms this chaotic data stream into discrete, meaningful commands (e.g., 5x-Inspect), inspired by the discrete objective lenses (e.g., 5x, 10x, 40x) on physical microscopes.
    \textbf{b} The recorder's novel data generation pipeline creates the Pathology-CoT dataset. For each captured action, an AI drafts a rationale explaining the expert's focus, which a pathologist then efficiently verifies or corrects in a human-in-the-loop workflow.
    \textbf{c} The \textbf{Pathology-CoT} dataset contains 5,222 conversational rounds from pathologists with diverse experience levels.
\textbf{d} The semi-automated, human-in-the-loop workflow is highly efficient, reducing labeling time by approximately six-fold compared to manual annotation from scratch.}
    \label{fig:overview}
\end{figure}
LLM-based agents have already begun to transform complex biomedical domains such as genetic experiment design and clinical diagnosis \cite{Huang2024CRISPRGPT,HealthAgentsClinical2024,BioDiscoveryAgent2024,CPathAgent2025}. In whole-slide image (WSI) analysis, an agentic paradigm is particularly well suited, since diagnosis is a dynamic, multi-step process rather than a static classification task. Here, an agent refers to a system capable of perceiving its environment, leveraging an LLM to perform reasoning and experiment orchestration, and executing a sequence of actions, such as zooming, panning, and selecting regions, to accomplish a complex diagnostic goal. Consequently, an agent that can systematically examine and analyze gigapixel WSIs provides a natural foundation for a collaborative AI partner for pathologists.
Although such model have a complex architecture, recent advances such as OpenAI o3~\cite{OpenAI2025O3SystemCard} and its variants~\cite{Zheng2025DeepEyes,Lai2025MiniO3} have already established capable agent frameworks with image zoom-in Chain-of-Thought (CoT) capability. In digital pathology, to ensure alignment with pathologists' routine operations and achieving scalability, the main bottleneck is the availability of high-quality, well-curated training data. Unlike programming or mathematical agents, clinical diagnosis cannot be guided by simple rule-based rewards. Agents instead need to learn how pathologists examine slides and which regions they attend to for each task \cite{Yu2021RLHealthcareSurvey,Ross2011DAgger}. While several datasets are now available, consisting of cropped image-caption pairs crawled from the internet~\cite{huang2023visual} or pairings of clinical WSIs and reports~\cite{ding2024titan}, these sources typically lack pathologist viewing behavior. Such lack of information underscores the urgent need to encode human knowledge and practice into reusable computational forms, which in turn requires direct expert involvement in creating such training data. However, gigapixel WSIs encompass diverse diagnostic scenarios and workflows \cite{Bankhead2017QuPath,Niazi2019DigitalPathologyReview}, often make dataset construction especially time-consuming, costly, and difficult to scale.

Modern digital pathology systems routinely record navigation events such as panning and zooming, thereby creating a continuous record of pathologist behavior \cite{Niazi2019DigitalPathologyReview}. 
Although these interaction logs can be produced at scale, such user behavior signals are both high-frequency and noisy, and thus require careful post-processing before they can serve as effective supervisory signals \cite{Mercan2016EyeTrackingPathology,Fawaz2019DLTimeSeries} (\textbf{Figure~\ref{fig:overview}a}). These ``digital exhaust'' traces can provide a scalable source of supervision for training human-aligned agents, but only if they are accurately captured, curated, and converted into structured, model-ready representations.

To address this gap, we introduce the \textbf{Pathology-CoT} framework, which comprises three key innovations. The \textbf{AI Session Recorder} segments continuous event streams and discretizes pathologists' behavior into a compact sequence of conversational text. These actions include exploratory scans at low or medium magnification and brief high-magnification inspections, each linked to LLM-friendly coordinates that define a Region of Interest (ROI). For each ROI, a VLM will generate a concise rationale and impression, which the pathologist can quickly verify or edit through a human-in-the-loop interface (\textbf{Figure ~\ref{fig:overview}b}). The resulting outputs collectively form the \textbf{Pathology-CoT dataset}, a high-quality behavioral corpus that captures not only where experts look but also why specific regions matter. This process distills meaningful behavioral signals from raw viewing logs, producing data ideally suited for training LLM-based agents. By integrating automated recording with expert validation, the AI Session Recorder substantially reduces annotation costs while maintaining clinical fidelity. In the gastrointestinal lymph node metastasis task within the Pathology-CoT dataset, this paradigm achieved a six-fold improvement in labeling efficiency.
To evaluate the effectiveness of behavioral data, we developed \textbf{Pathology-o3}, an agent trained on recorded viewing behavior in Pathology-CoT. The model achieved 84.5\% precision, 100.0\% recall, and 75.4\% accuracy, outperforming the state-of-the-art closed-source commercial thinking-with-image OpenAI o3 model, which reached 46.7\% precision, 87.5\% recall, and 57.8\% accuracy. Across different backbones, incorporating behavioral guidance increased both precision and recall by an average of 11.8\% and 17.9\%, respectively. Furthermore, the agent's learned viewing policy demonstrated strong generalization, maintaining robust performance (97.6\% recall) on a challenging independent external validation cohort. Critically, Pathology-o3's navigation strategy was highly aligned with that of a senior attending pathologist—both qualitatively and quantitatively—a stark contrast to the scattered, low-yield navigation of other VLMs. 

With these contributions, this study tackle the central bottleneck for innovating and advancing digital pathology agents: obtaining data that is both clinically meaningful and scalable. By converting routine viewing logs into task-conditioned behavioral supervision paired with low-cost, expert-verified reasoning, we supply the supervision required for existing architectures to evolve into capable, human-aligned agents. The Pathology-CoT framework thus represents a pivotal step toward behavior-grounded, clinically trustworthy AI systems in computational pathology.


\section{Results}

\subsection{A scalable data engine converts viewing behavior into agent-ready supervision}

Diagnosing a whole-slide image for conditions like lymph node metastasis is a dynamic process of visual inquiry. 
Pathologists begin with a low-power overview, zoom into suspicious areas for high-power analysis, and iteratively navigate across the gigapixel canvas to build a conclusion (\textbf{Figure~\ref{fig:overview}a} left). 
This sequence of interactions is highly goal-oriented: the keyboard and mouse movements embody the pathologist's tacit knowledge, reflecting an expert's experience-driven strategy for searching for evidence. 
While modern digital pathology systems routinely capture this behavior in interaction logs, the raw data exists as a high-frequency, noisy, and enormously long stream. 
Such ``digital exhaust'' is too complex and massive for direct use in training machine learning models, as the chaotic input exceeds the context limits of current AI and prevents effective learning.
This gap means that the vast expertise exercised in daily clinical practice remains in a format that is not amenable to model training. 
A systematic methodology to record, structure, and encode this knowledge is fundamentally required to train sophisticated AI agents that can serve as true diagnostic co-pilots or autopilots.

Given this urgent challenge, we developed the pathology ``\textbf{AI Session Recorder}'', a tool that aims at resolving the data bottleneck of training next-generation medical AI agent by collecting pathologist's behavior data at scale. AI Session Recorder performs two key functions to transform unusable digital pathology image inspection raw logs into structured, agent-ready supervision (\textbf{Figure~\ref{fig:overview}a}).
First, it automatically analyzes the complex viewing process to identify and discretize the most important areas of focus (\textbf{Figure~\ref{fig:overview}b}). 
Second, inspired by the discrete objective lenses of a physical microscope, this process converts the chaotic event stream into a compact sequence of meaningful \emph{behavioral commands}. 
Each command is defined as either a broad, low-magnification \texttt{<inspect>} or a rapid, high-magnification \texttt{<peek>}, and is paired with a standardized region-of-interest (ROI) box. 
The recorder also captures clinical reasoning by using these expert‑identified ROIs to prompt a vision‑language model (VLM), which drafts a rationale explaining why the region was examined and what key findings are present (\textbf{Figure~\ref{fig:overview}b}, left). The model‑generated rationales are reviewed and, when necessary, edited by two pathologists, then consolidated into the Pathology‑CoT agent‑training dataset (\textbf{Figure~\ref{fig:overview}b}, right, shows an example).

To implement and verify this AI Session Recorder, we focused on a clinically representative task, namely N-staging of colorectal cancer (CRC) lymph-node metastasis, which is both common and labor-intensive. We collected behavioral data from eight pathologists at Stanford Medicine using a standard viewer nuclei.io \cite{huang2025pathologist} that recorded all navigation events of making diagnosis for adjacent lymph node metastasis in colorectal cancer. Such raw data contains 25 cases, 137 slides across 8 pathologists (four attending pathologists, two fellows, and two residents) with a total of 10.6 hours of inspection (\textbf{Figure~\ref{fig:overview}c}).
The raw logs were noisy and high-frequency, with active navigation captured at approximately 10 Hz and averaging more than 257 distinct viewport events per slide (\textbf{Figure~\ref{fig:pipeline_detail}}). Using this stream directly as supervision was infeasible because feeding the raw viewports into a VLM would produce over 500K visual tokens per case, far exceeding the current context limits and could also prevent effective learning. In contrast, the AI Session Recorder reduces excessive context length by filtering noisy, non-informative behavior data, while preserving the fidelity of pathologists’ behavioral signals. AI Session Recorder is also robust, showing similar behavior pattern across same pathologist's two viewing session (\textbf{Figure \ref{fig:session_compare}}). 
Using this tool, human pathologists can rapidly verify AI-drafted behavior and rationale by accepting, revising, or rejecting them.

In this verification setting, we invited two additional pathologists (C.H. from University of California at San Francisco and S.K. from the Hospital of the University of Pennsylvania) to review the chain-of-thought data derived from the viewing logs of the eight pathologists, including proposed regions to zoom, per-ROI rationales (why to zoom and what is seen), the synthesized summary, and the final impression (\textbf{Figure \ref{fig:human_in_loop_review_software}}). During verification stage, two additional pathologists can accept the generated text, edit it, or indicate that a proposed behavior was unnecessary (``no zoom here''). Each zoom-in constitutes a two-round exchange: a decision to zoom and a description of the findings. In a two-user timing study (\textbf{Figure~\ref{fig:overview}d}), verification took 12.1 s and 9.7 s per round for Users 1 and 2, respectively. When edits were needed, revisions took 55.5 s and 40.2 s, respectively. Averaged across rounds, the verify-and-edit workflow required 18.8 s (User 1) and 16.3 s (User 2). For comparison, ``annotation from scratch'' measures the total time to decide where to zoom and to explain (by typing or dictation) why to zoom and what is found; this required 106.2 s and 85.2 s by typing or 56.7 s and 64.9 s by verbal dictation. Based on these observations, we conclude that our semi-automated, human-in-the-loop process is roughly 5--6\(\times\) faster than typing and 3--4\(\times\) faster than dictation, with over 80\% of AI-generated text requiring no edits, making the AI Session Recorder a low-cost and scalable data collection engine. See \textbf{Appendix} for details on the software and experiment implementation.

Based on the manual filtering, the resulting \textbf{Pathology-CoT dataset} is a structured collection of expert-validated, task-conditioned ``behavior + reasoning'' pairs. Pathology-CoT dataset is a large-scale public dataset derived from 10.6 hours of recorded interactions from eight pathologists with diverse experience levels (attendings, fellows, and residents) across 921 sessions (\textbf{Figure~\ref{fig:overview}c}), the dataset contains 5,222 conversational rounds of agentic vision CoT training data. The reasoning is detailed: broad, low-magnification \texttt{<inspect>} actions are accompanied by an average of 152 words of description, while the analysis of fine cellular-level findings from high-magnification \texttt{<peek>} actions averages 82 words. This rich dataset captures not only \emph{where} and \emph{how} an expert looks but also embeds the clinical reasoning for \emph{why}, providing a robust, multi-modal foundation for training and evaluating pathology agents. Unlike PathChat+~\cite{Chen2025PathChatPlus}, whose reasoning was based on VLM-generated text, our dataset captures the thinking process of expert pathologists. To our knowledge, this is the first chain-of-thought pathology behavior data that captured and established from expert pathologists.

\begin{figure}[hbtp]
    \centering
    \includegraphics[width=0.8\linewidth]{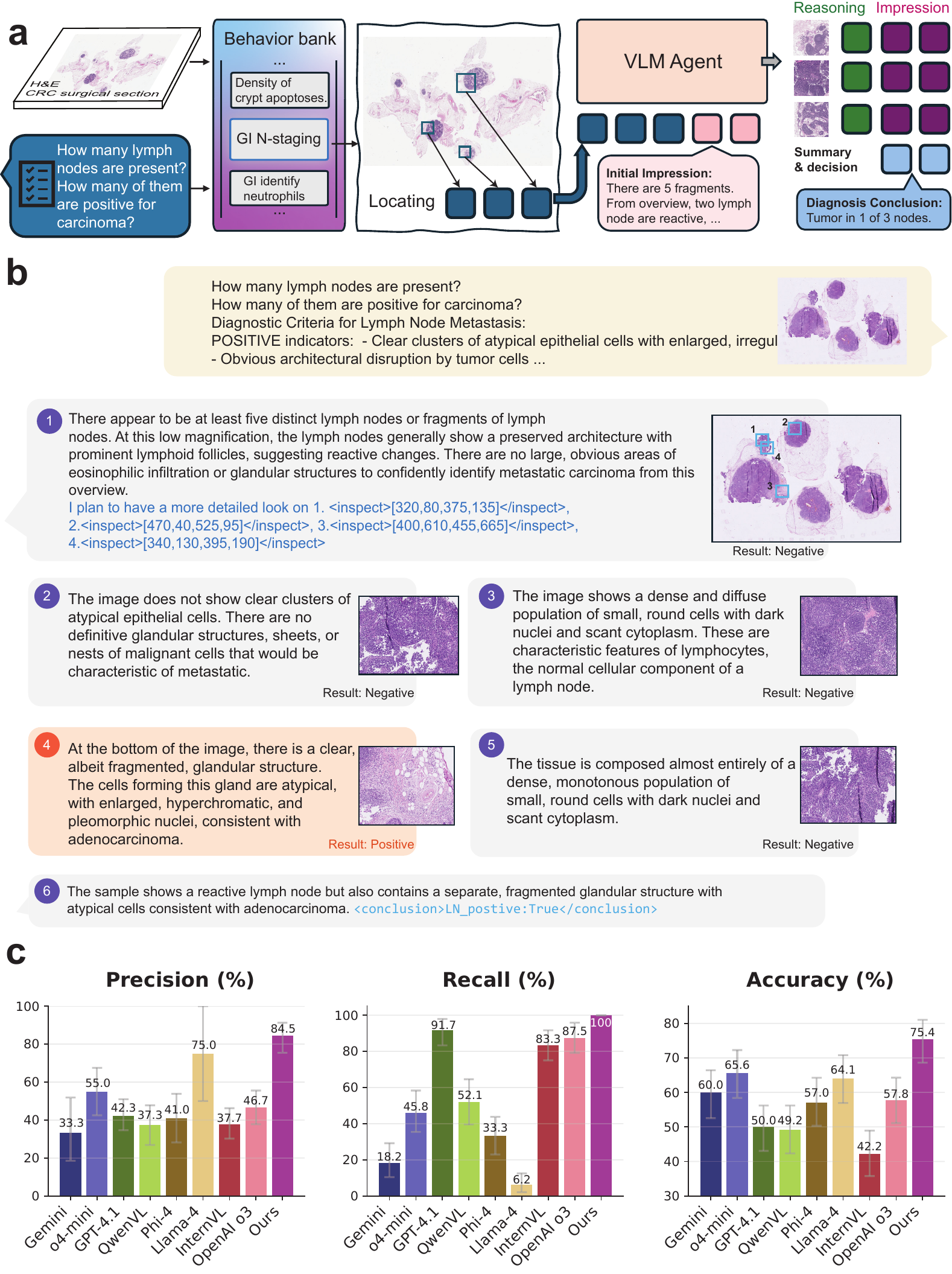}
    \caption{Overview of \textbf{Pathology-o3} and its performance. 
    \textbf{a} Workflow: a task prompt initializes the agent, which proposes candidate regions using a behavior bank/locating module; cropped ROIs are sent to a VLM that generates per‑ROI reasoning, then summarized into an impression and final diagnosis. 
    \textbf{b} Example ``thinking with image” on one slide. The agent lists planned inspections, provides step‑wise descriptions for each ROI (blue boxes), and synthesizes a case‑level conclusion. Positive and negative findings are indicated for each step. 
    \textbf{c} Quantitative comparison on the lymph‑node metastasis task. Pathology-o3 achieves the best overall accuracy with perfect or near‑perfect recall, outperforming strong VLM baselines. Error bars represent 95\% bootstrap confidence intervals.}
    \label{fig:asset3}
\end{figure}

\subsection{Behavior-guided supervision enables accurate pathology agents}

AI Session Recorder enables a scalable way to record tacit diagnostic knowledge from pathologists at scale. By that we are able to collect a scalable ``Pathology-CoT dataset'' that is ready for machine learning. Based on such format of dataset, it allows (i) digitize and store pathologists' behavior; (ii) guiding machine learning model to locate important regions from a WSI; and (iii) training an agentic AI system to make clinical diagnosis in the same manner as pathologists.

To demonstrate the utility of that, we designed a task-conditioned agent ``\textbf{Pathology-o3}'' that mimics a pathologist's workflow. Learned from the training data Pathology-CoT in example of lymph node metastasis, the agent first decides where to look, then analyzes what it sees (\textbf{Figure~\ref{fig:asset3}a}). To emulate this first step, it uses a task-specific \emph{Behavior Predictor} trained on the complete set of discretized behavioral commands from aforementioned Pathology-CoT dataset, then learning to propose diagnostic ROIs on a new slide in our internal and external validation datasets. Once these ROIs are identified, the system performs \emph{behavior-guided thinking}: for each region, it auto-crops a high-resolution field, injects task context into the prompt (case-level goal and information, and ROI-level intent), and then prompts a VLM to generate per-ROI reasoning, which is subsequently synthesized into a case-level summary. This capability by combining a learned human viewing policy with a powerful reasoning AI, allows the agent to produce a step-by-step visual chain-of-thought for a given WSI, demonstrating its ability to systematically analyze a slide in a human-aligned, easy to understand and validate manner (\textbf{Figure~\ref{fig:asset3}b}). 

We evaluated our framework against a panel of state-of-the-art VLMs.
On the lymph-node metastasis benchmark, our behavior-guided agent achieved \textbf{84.5\% precision}, \textbf{100\% recall}, and \textbf{75.4\% accuracy} (\textbf{Figure~\ref{fig:asset3}c}). The next-best model, OpenAI o3, reached 46.7\% precision, 87.5\% recall, and 57.8\% accuracy. VLMs without chain-of-thought ability performed notably worse and exhibited imbalanced trade-offs: for example, Phi-4 achieved high recall (91.7\%) but low precision (42.3\%), while Llama-4 showed higher precision (75.0\%) but extremely low recall (6.2\%). Across the remaining baselines, precision ranged from 33.3--65.6\%, recall from 18.2--91.7\%, and accuracy from 42.2--65.6\%.
These results demonstrate the importance of grounding reasoning in expert visual search patterns; for nuanced diagnostic challenges like lymph node metastasis, generic reasoning is insufficient.
Performance hinges on first learning where and how to look (i.e., thinking with image), which our behavior-aligned data supplies. More detail on the setting for VLM and performance metrics are shown in \textbf{Table \ref{tab:vlm_comparison}} and \textbf{Table \ref{tab:fig2_model_performance}}.
It is important to note that multiple instance learning (MIL) and related approaches require scanning every patch of a whole-slide image, making the process computationally intensive and fundamentally different from the goal-oriented behavior of human experts. Because such exhaustive patch-level processing cannot be efficiently guided or learned from human behavioral data, these MIL-based methods were not included in the comparison.

\subsection{Pathology-o3 and learned behavior can generalize on external validation dataset}
\begin{figure}[hbtp]
    \centering
    \includegraphics[width=0.85\linewidth]{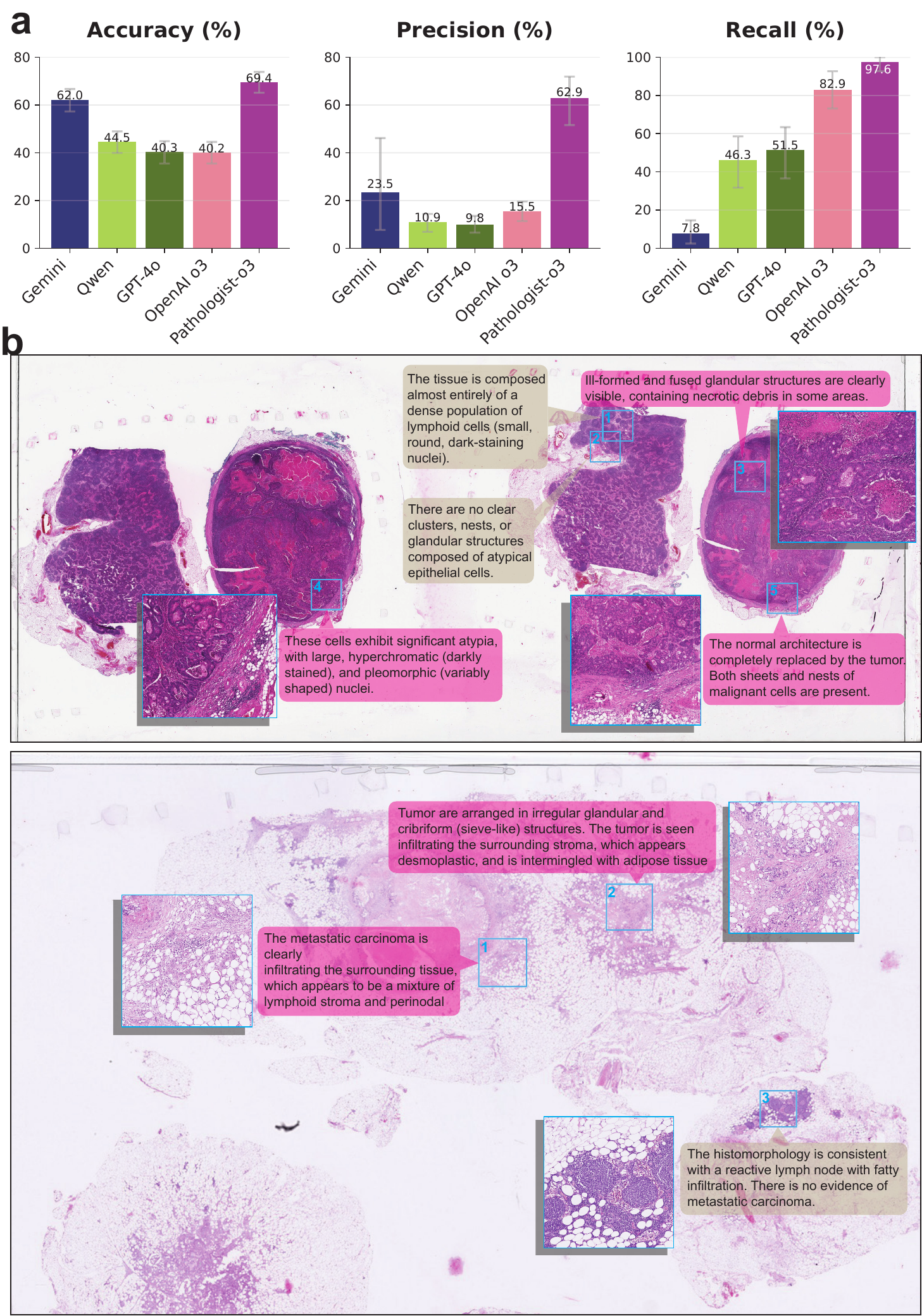}
    \caption{Quantitative and qualitative results on the external LNCO2 validation cohort. \textbf{a} Bar charts comparing Accuracy, Precision, and Recall of Pathology-CoT with other VLM backbones. Error bars represent 95\% bootstrap confidence intervals. \textbf{b} Qualitative examples of the model's output on two slide images, with highlighted regions and corresponding textual descriptions.}
    \label{fig:asset4}
\end{figure}

To test for generalization, Pathology-o3 and the Behavior Predictor, trained on Stanford Medicine data, was evaluated on a completely independent external colon adenocarcinoma lymph node metastasis dataset: LNCO2~\cite{maras_2020_lnco2}. This challenging cohort consists of 321 whole-slide images from Sweden, introducing domain shifts from different scanners and magnifications. Despite these variations, Pathology-o3 trained on United State pathologist behavior maintained robust performance, substantially outperforming other leading VLMs on this benchmark (\textbf{Figure \ref{fig:asset4}a}).

Quantitatively, our Pathology-o3 agent AI achieved an accuracy of 69.4\%, a precision of 62.9\%, and a near-perfect recall of 97.6\%. These results demonstrate a significant leap in performance. For instance, Pathology-o3's precision was more than double that of the next-best model (Gemini at 23.5\%), and its exceptional recall highlights a strong ability to reliably identify positive cases. More detail on performance metrics are shown in \textbf{Table \ref{tab:external_validation_performance}}.

For example, as shown in \textbf{Figure \ref{fig:asset4}b}, in the first case (inspects 1 and 2; highlighted in blue squares) and the second case (inspect 3), the Behavior Predictor identified tinctorial differences and architectural disruption. For instance, in inspect 2, Behavior Predictor suggests ``There are no clear clusters, nests, or glandular structures composed of atypical epithelial cells.'' Paler clusters were detected that interrupted the normal diffuse pattern of the lymph-node parenchyma, appearing as disorganized aggregates or nodules. In the first case (inspects 4 and 5) and the second case (inspect 1), the Behavior Predictor located the lymph-node capsule edge with tinctorial differences and correctly determined that the node was positive for tumor metastasis. These findings suggest that supervision at the level of how and where to look yields a transferable viewing policy that is more robust to domain shifts in image appearance than direct pixel-level supervision.

\begin{figure}[hbtp]
    \centering
    \includegraphics[width=1\linewidth]{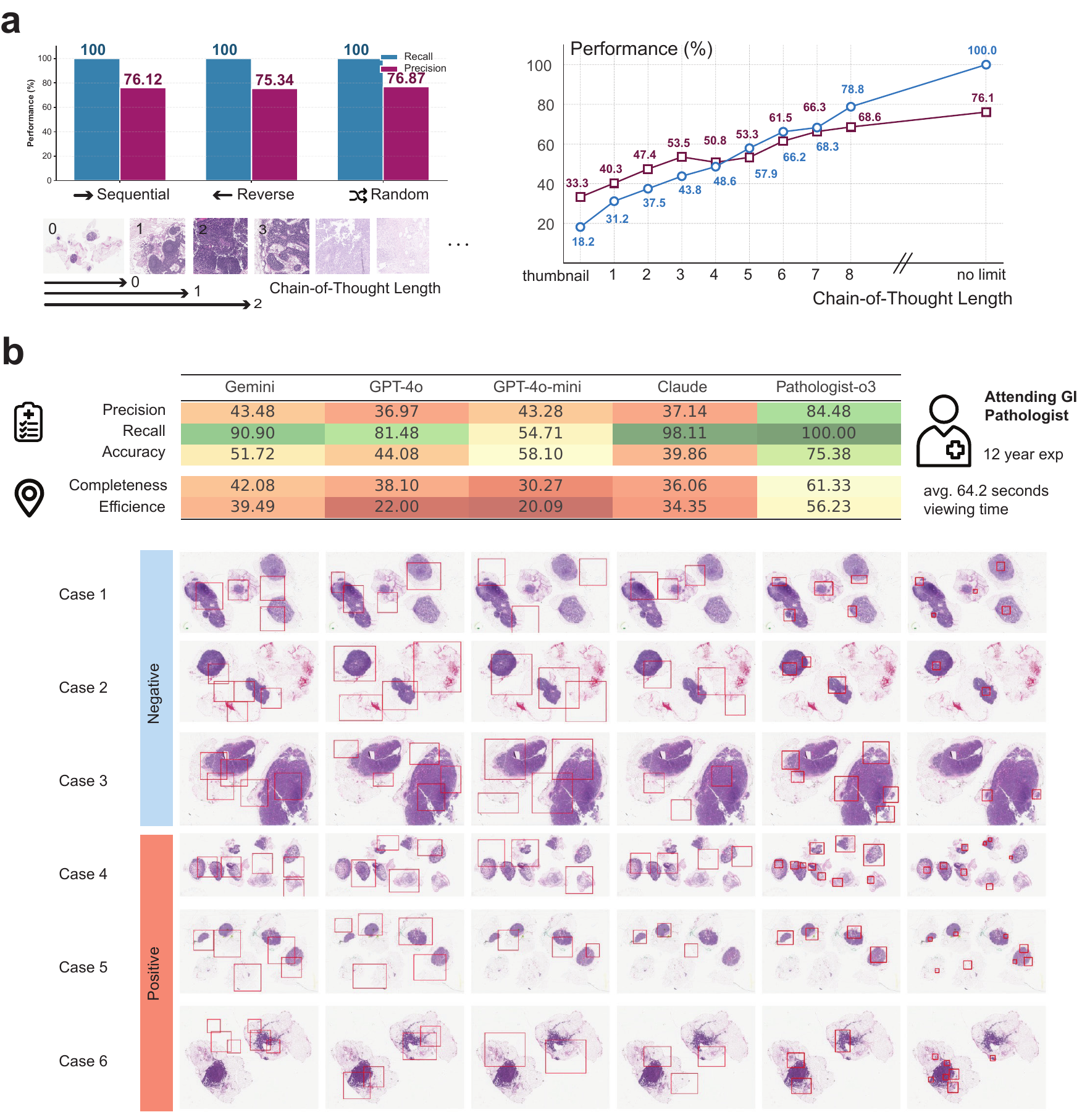}
    \caption{Analysis of reasoning length and order. \textbf{a} The effect of sequential, reverse, and random order of evidence, and the impact of chain-of-thought length on performance. \textbf{b} Qualitative and quantitative comparison of different models' predictions against a senior pathologist's behavior.}
    \label{fig:asset8}
\end{figure}
\subsection{Pathology-o3 learns where to look}
\label{sec:compare_to_llm_generated_behavior}
We next examined two key properties of our behavior-guided agent: the robustness of its learned policy and the fundamental importance of navigation itself. 
We first evaluated robustness by permuting the order of ROI presentation (forward, reverse, and random). The agent's performance varied only minimally across these settings (\textbf{Figure~\ref{fig:asset8}a}, left), indicating that it synthesizes evidence effectively without dependence on a fixed sequence.
This suggests that when identify lymph node metastasis, as long as diagnostically relevant regions are identified, the different viewing habits of pathologists can be learned and successfully deployed within an agentic system.

Second, we analyzed the agent's performance when we capped the maximum number of ROIs it could inspect, analogous to CoT length. 
As shown in \textbf{Figure~\ref{fig:asset8}a} (right), performance improved steadily as thinking process getting longer, with notable gains in recall while preserving high precision. This finding demonstrates that when provided with a sequence of diagnostically relevant regions, VLMs are highly effective at synthesizing the evidence to reach an accurate conclusion.

This proficiency in synthesizing evidence, however, is entirely contingent upon the VLM being provided with diagnostically relevant regions. This allowed us to directly test our central hypothesis: that the ability to navigate and locate these regions is a distinct, learned skill derived from expert viewing behavior—a skill for which current VLMs lack the necessary training data. To investigate this, we evaluated VLM out-of-the-box navigation capabilities. Rather than being guided by our Behavior Predictor, we let VLMs to select their own ROIs. We then compared their navigation against a senior attending pathologist's path using two metrics (described in \textbf{Methods}): completeness, which represents the fraction of expert-viewed regions recovered; and efficiency, which represents the proportion of selected regions that were clinically relevant. As shown in \textbf{Figure~\ref{fig:asset8}b} (upper), unguided VLMs performed poorly on both measures. Qualitatively, their ROI selections were scattered and low-yield, whereas Pathology-o3 concentrated on clinically salient sites that closely overlapped with the an attending pathologist’s real behavior (\textbf{Figure~\ref{fig:asset8}b}, lower). These results reveal a fundamental gap between analysis and navigation. The strong analytical ability of VLMs reflects training on large corpora of static, pre-cropped images, whereas their failure to navigate stems from the absence of data capturing continuous, visually guided interaction with complex environments. This finding reinforces our data-centric approach: building effective pathology agents requires digitizing the tacit, procedural knowledge of expert navigation into a structured format suitable for learning.


\subsection{Practicality, scalability, and the value of learned behavior}
\begin{figure}[hbtp]
    \centering
    \includegraphics[width=1\linewidth]{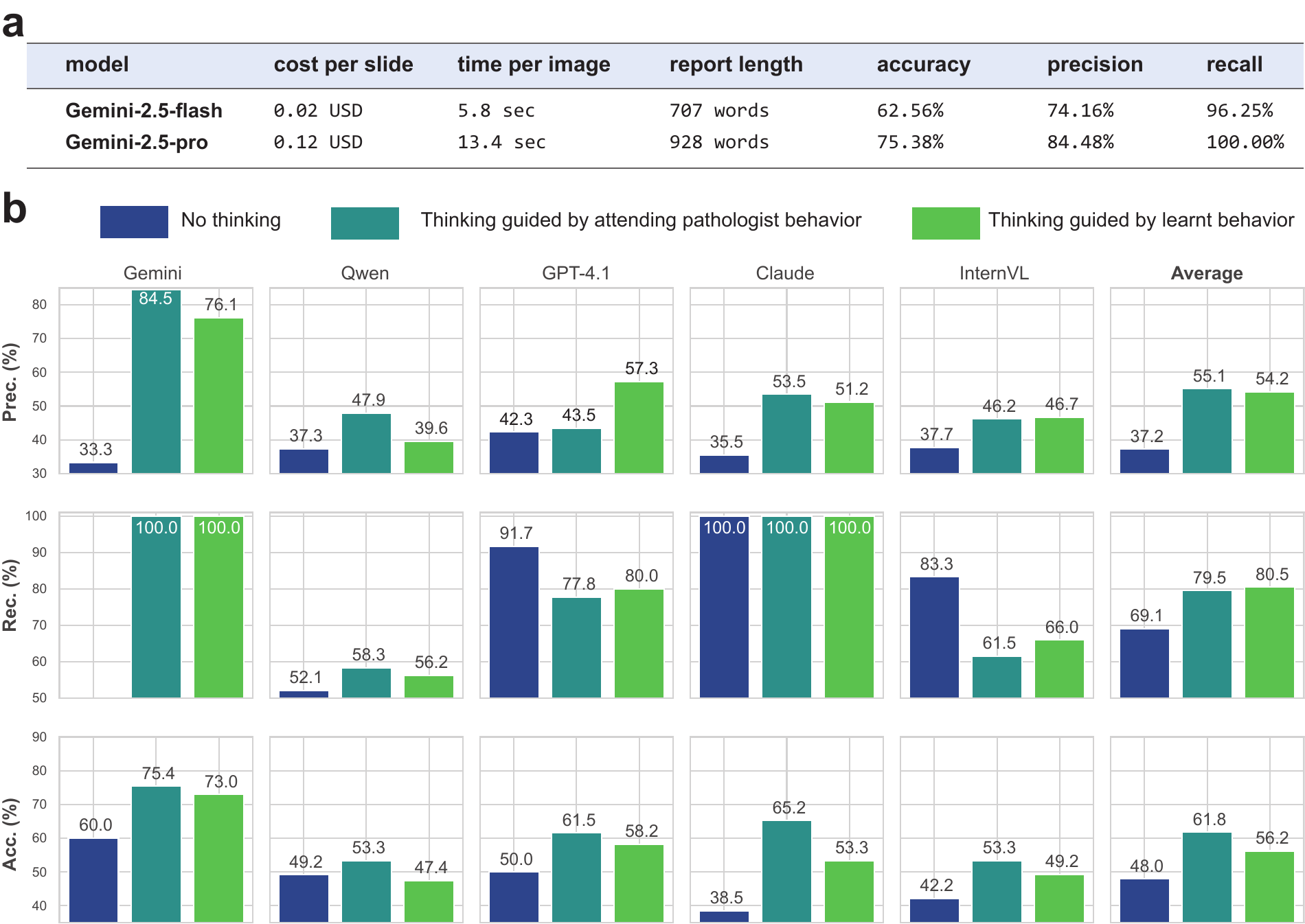}
    \caption{Comparison of different ``thinking'' modes and cost-benefit analysis. \textbf{a} A table comparing cost, time, and performance between high-cost reasoning model Gemini-2.5-pro and low cost. \textbf{b} Performance comparison of ``Non-agent'', ``Agent guided by real behavior'', and ``Agent guided by learned behavior'' modes across various VLMs.}
    \label{fig:asset9}
\end{figure}

Finally, we analyzed the practical viability and scalability of Pathology-o3. \textbf{Figure~\ref{fig:asset9}a} provides a cost and latency snapshot for a representative WSI analysis, comparing a large, powerful model Gemini-2.5-pro with a smaller, faster one Gemini-2.5-flash. The results demonstrate that Pathology-o3 is already feasible with current backbones. 
More importantly, our framework is modular by design, allowing the core reasoning engine (the VLM) to be upgraded as lastest foundation models evolve. 
This design ensures the system is future-proof: it can readily incorporate next-generation models to become progressively faster, cheaper, and more capable over time.

Next, we investigated the value of the behavior-aligned data itself. We compared three conditions across multiple VLM backbones (\textbf{Figure~\ref{fig:asset9}b}): (i) Non-agent is a baseline without guidance and diagnosis with only look at the WSI thumbnail once (similar to PathChat+~\cite{Chen2025PathChatPlus} approach); (ii) Pathology-o3 guided by \emph{real} behavior from a senior attending GI pathologist (a practical upper bound) as we shown in \textbf{Figure~\ref{fig:asset8}b}; and (iii) Pathology-o3 guided by \emph{learned} behavior from the behavior predictor. The addition of behavioral guidance yielded a profound improvement in performance. Compared to the baseline, guidance from the learned viewing policy increased precision by 11.8\% and recall by 17.9\%. The gains from real expert behavior were similarly large, boosting precision by 11.9\% and recall by 19.1\%. This translated to a marked increase in overall accuracy, improving it by 9.2\% and 7.1\%, respectively. Critically, Pathology-o3's \emph{learned} behavior closes most of the performance gap to \emph{real} behavior. This has a powerful implication for scalability: instead of requiring continuous, large-scale expert data collection, we can train a compact and efficient behavior predictor on a limited dataset. This predictor effectively distills the expert's viewing policy, creating a scalable asset that can generate human-aligned guidance for any VLM, reinforcing our central hypothesis that the most effective path to a capable pathology agent is by first building the right data engine.

\section{Discussion}
The rapid maturation of vision–language models is creating a new frontier for artificial intelligence in medicine, shifting the paradigm from task-specific classifiers to versatile, reasoning-based agents. Digital pathology, characterized by gigapixel images and multi-step diagnostic workflows, is especially suited to an agentic approach that can automate diagnostic reasoning and accelerate clinical and scientific discovery. However, the primary obstacle to developing capable pathology agents has shifted from modeling limitations to the availability of high-quality data—specifically, supervision that is not only large-scale but also deeply aligned with the procedural and tacit knowledge of clinical experts.

The central challenge is that this expert knowledge is not readily available in a structured format. While modern pathology viewers generate vast digital logs of user interactions, this raw behavioral data is high-frequency, noisy, and enormously long. It captures the outcome of an expert's thought process but not the discrete, intention-driven steps that are amenable to model training. This results in a fundamental ``analysis–navigation gap'': while VLMs are trained on vast corpora of static images and are thus adept at analyzing pre-selected regions, they are not trained on the dynamic, interactive data needed to learn the procedural skill of navigating a slide to find those regions in the first place.

To bridge this gap, we developed the \textbf{AI Session Recorder}, a data engine that digitizes the tacit knowledge of expert navigation into our \textbf{Pathology-CoT dataset}. We demonstrated the value of this behavior-aligned data by building a simple agent AI, \textbf{Pathology-o3}, as a controlled testbed. Our results show that this data provides a universal, model-agnostic performance multiplier, significantly improving the diagnostic accuracy of a wide range of VLM backbones. We found that the learned viewing policy is robust, generalizing effectively to an entirely different hospital system with different scanners and patient populations. Further analysis revealed that this policy is not only robust to variations in viewing habits but is also essential for effective diagnosis, as leading VLMs without this guidance fail at the fundamental task of navigation. Finally, we concluded that this data-centric approach is both practical and scalable; the expert viewing policy can be distilled into a compact, efficient model that provides nearly all the benefits of real expert guidance, creating a powerful asset for building the next generation of capable, human-aligned pathology agents.

Our approach is fundamentally different from other pathology AI agent methods like PathChat+~\cite{Chen2025PathChatPlus}, which delegates the selection of where to zoom and why to an VLM ``supervisor'' such as GPT-4o. This strategy is limited because the procedural viewing policy of a pathologist is tacit, experience-based knowledge that is not documented in textbooks or on the internet, making it difficult for VLMs to acquire.
Consequently, the navigation paths generated by the VLM supervisor often diverge from expert trajectories and lack diagnostic utility, as we illustrate in \textbf{Sec \ref{sec:compare_to_llm_generated_behavior}}. 
In contrast, by directly learning from recorded expert navigation and pairing it with pathologist-verified rationales, our method provides explicit supervision for both the ``where'' and the ``why,'' enabling truly human-aligned behavior and stronger performance.

While this study introduced the powerful ``Pathology-CoT'' framework, this work still has several limitations, primarily centered on our data-centric methodology. First, our findings on viewing policy, such as robustness to viewing order, are specific to the "search-and-find" task of metastasis detection. This may not generalize to tasks requiring a systematic, sequential evaluation, like architectural pattern grading in prostate cancer or counting mitoses along a tumor boundary, where the viewing path itself is a critical component of the diagnosis. To begin addressing this, we are continuously expanding our public dataset with behavior logs from a wider range of pathological tasks, with ongoing data releases available on our project website. Second, our efficient human-in-the-loop workflow for generating reasoning labels introduces a potential for AI anchoring bias, where the initial AI-generated text could subtly influence the expert's final revision. Finally, a practical challenge lies in the data source itself. Unlike standardized web-scale corpora, pathologist viewing logs are heterogeneous, with different software platforms using varied and often unstructured formats. Tapping into this data requires an initial engineering effort for each system. However, the significant advantage of our approach is that once this integration is complete, data collection becomes entirely passive, imposing no additional burden on the pathologists' routine clinical workflow.

The insights from this study were enabled by our \textbf{AI Session Recorder}, the data engine we developed to curate \textbf{Pathology-CoT}, the first large-scale public dataset of expert pathologist viewing behavior. Using this engine, we are continuously expanding the Pathology-CoT dataset to cover a wider range of tasks. We anticipate that our open-source methodology and this growing agentic data resource will benefit the medical AI community by providing a new modality of supervision—procedural behavior—and by offering a blueprint for creating similar data engines in other clinical domains where expert interaction is the key to building capable, human-aligned AI.

\newpage
\section{Method}

\subsection{Data Collection: Pathologist Behavior}
\label{subsec:data_collection}

This study was approved by the Institutional Review Board (IRB \#857258) of the Perelman School of Medicine at the University of Pennsylvania. We collected interaction data from eight pathologists with varying experience levels (four attendings, two fellows, and two residents)  analyzing 137 diagnostic Whole Slide Images (WSIs) from 25 cases of colorectal cancer (CRC) lymph node (LN) metastasis. The data collection setup, using the nuclei.io software, recorded the following:
\begin{itemize}
    \item \textbf{High-Resolution WSIs:} The original gigapixel images, which were digitized using a Leica Aperio Scanner at 40x magnification (0.25 µm per pixel).
    \item \textbf{Timestamped Viewports and Navigation Actions:} The sequence of user activities was recorded into a log file with timestamps detailed in milliseconds. These logs captured the Field of View (FoV) coordinates and zoom level changes as pathologists navigated the WSIs.
    \item \textbf{Diagnostic Result:} The final annotations marking positive LNs were recorded for each WSI, which formed the basis for the final diagnostic findings.
\end{itemize}
This raw data captures the dynamic, multi-modal behavior of expert pathologists during routine diagnostic tasks (\textbf{Figure~\ref{fig:dataset_creation}}, left panel).

\subsection{AI Session Recorder: Discretizing Behavior to Extract ROIs}
\label{subsec:behavior_preprocessing}

The foundation of our approach is the AI Session Recorder, a computational pipeline designed to transform the raw interaction data gathered from pathologists—a high-frequency, continuous, and inherently noisy stream of timestamped viewports (\textbf{Figure \ref{fig:pipeline_detail}}) into structured, agent-ready data. Pathologists' navigation is rapid and complex; a single slide analysis can generate an average of 257.3 viewport events. Directly using this raw data for model training would be chaotic and impractical. If each of these ~257 viewports were treated as a separate 1024x1024 patch, it would generate over 500K tokens from the visual modality alone. This volume is prohibitive for prominent large models, and could severely degrade performance.

The AI Session Recorder processes this raw stream using a series of heuristics inspired by the physical use of a microscope, which has discrete, not continuous, magnification levels. We analyze the temporal patterns of viewport changes, as shown in \textbf{Figure \ref{fig:pipeline_detail}}, to segment the continuous stream into discrete actions. We define two primary types:
\begin{itemize}
\item \textbf{<inspect>:} This action represents a broad, exploratory examination. It is identified when a pathologist either holds their viewport static over an area for more than one second or continuously pans across a region for more than two seconds.
\item \textbf{<peek>:} This action captures a quick, high-magnification look at fine cytological details. It is implemented by identifying a rapid zoom to the WSI's native resolution and capturing the central 1024x1024 pixel area.
\end{itemize}

These initial actions are then filtered to remove low-magnification overviews, merged if they have high spatial overlap (IoU > 0.8), and pruned to prioritize the most specific, high-magnification views. Finally, each resulting action's magnification level is binned to the nearest standard microscope objective, and its bounding box is normalized to a standard size. For instance, a 12x view becomes a <10x-inspect> action. As shown in \textbf{Figure \ref{fig:pipeline_detail}}, this converts the raw data into a clean, compact sequence of behavioral commands. This abstracted action sequence, composed of a command type and a standardized ROI box, serves as the basis for both training our agent and generating reasoning labels.

\subsection{AI Session Recorder: Rationale Generation}
\label{subsec:gt_generation}

After extracting behavioral ROIs, the second function of the AI Session Recorder is to generate a corresponding clinical rationale for each one. This is achieved through a scalable, semi-automated human-in-the-loop workflow that systematically generates the ground truth components for our agent, specifically the reasoning (think) and findings (answer) behind each action. The process unfolds in three stages:

\begin{enumerate}
\item \textbf{Contextual Prompting:} For each ROI, a powerful Vision-Language Model (VLM), such as Gemini 2.5 Pro, is prompted with rich context. This includes the overall task description, the low-magnification image of the ROI, a high-magnification crop from its center, and, when available, snippets of the pathologist's transcribed verbal commentary during the original examination.

\item \textbf{Guided Generation:} The VLM is instructed to synthesize a concise rationale that logically explains both \textbf{why} an expert would focus on this region (the thinking process) and \textbf{what} the key morphological findings are within it (the answer).

\item \textbf{Pathologist Review and Correction:} Critically, the AI-generated draft is presented to a human pathologist in a lightweight interface for rapid validation, as shown in \textbf{Figure~\ref{fig:human_in_loop_review_software}}. The interface displays the relevant thumbnail, the zoomed-in ROI, and an optional cytology-level crop alongside the generated text for "Thumbnail Impression," "Why Zoom," and "Findings." The pathologist can then efficiently accept, edit, or reject the text to ensure its clinical accuracy and alignment with their actual reasoning process.
\end{enumerate}

This "review-over-authoring" paradigm proved to be 8.6 times faster than manual annotation from scratch. It enables the efficient creation of a high-quality dataset composed of paired "behavior + reasoning" data, which is crucial for both training our agent and generating its reasoning labels.

\setcounter{table}{0}
\renewcommand{\thetable}{S\arabic{table}}

\subsection{The Pathology-CoT Dataset}
The curated rationale and impression components are structured alongside the corresponding state information to form the final Pathology-CoT dataset. As illustrated in (\textbf{Figure~\ref{fig:dataset_creation}, right}), the data can be formatted into sequences of state-action tuples, or arranged in a conversational format suitable for prompting VLMs using In-Context Learning.

The final output of our pipeline is the \textbf{Pathology-CoT dataset}, a structured collection of expert-validated, agent-ready training data. The dataset is derived from 10.6 hours of recorded interactions from eight pathologists at Stanford Hospital (including attendings, fellows, and residents) analyzing WSIs for the task of colorectal cancer lymph node metastasis. The current version contains 921 sessions and 5,222 conversational rounds. Each round consists of a behavioral command, a standardized ROI, and the corresponding expert-validated reasoning. The reasoning is richly detailed, with broad \texttt{<inspect>} actions accompanied by an average of 152 words of description, and high-magnification \texttt{<peek>} actions averaging 82 words on cellular-level findings. We also show the detailed distribution of the dataset using sankey plot as shown in \textbf{Figure \ref{fig:sankey_analyze}}.

\subsection{Pathology-o3 Agent Framework and Evaluation}

Inspired by recent advancements in decoupling complex vision-language tasks with tool-driven visual exploration \cite{su2025thinking}, we conceptualize WSI diagnosis as a sequential, multi-stage process. 
For a given WSI and a high-level objective (e.g., ``Find all positive lymph nodes"), our pipeline operates in three key stages:

\begin{enumerate}
    \item \textbf{Global Thumbnail Analysis:} The system first processes a downsampled thumbnail of the WSI to gain global context and form an initial impression.

    \item \textbf{Behavior Prediction:} In response to the given task, a Large Language Model (LLM) functions as a router, selecting the most suitable behavior predictor from a predefined ``behavior bank.'' 
    This predictor then proposes a set of candidate Regions of Interest (ROIs) as we will detail in next section. 
    These ROIs, represented as bounding boxes, act as ``Location Prompts'' that direct the system's focus to specific areas for closer inspection, mimicking how a human pathologist identifies salient regions for magnification.

    \item \textbf{Reasoned Analysis and Summarization:} Each candidate ROI is subsequently cropped at a higher magnification and processed by a central Vision-Language Model (VLM), termed the ``Reasoning Module.'' This module generates an explicit reasoning chain and detailed findings for each region. 
    After analyzing all ROIs, the module synthesizes this information to produce a final diagnostic summary. 
    Notably, unlike Pathology-CoT, Pathology-o3 does not require the VLM to rationalize the selection of an ROI, as this task is handled exclusively by the behavior predictor.

\end{enumerate}
This decoupled pipeline enables the system to efficiently scan a large WSI and concentrate its advanced reasoning capabilities on the most pertinent areas, thereby mirroring an expert pathologist's workflow. To determine the optimal VLM for the Global Thumbnail Analysis, Reasoning Module and Summarization, we conducted a comparative analysis detailed in \textbf{Table~\ref{tab:vlm_comparison}}. Gemini-2.5-pro demonstrated the best overall performance and was subsequently used in all of our experiments.

\begin{table}[hbtp]  
    \centering
    \caption{Performance comparison using different VLMs in Pathology-o3 (Behavior Predictor Fixed). Best performed method(s) were highlighted in bold font.}
    \label{tab:vlm_comparison}  
    \begin{tabular}{lccc} 
        \toprule
        \multirow{2}{*}{Model (Core VLM Backbone)} & \multicolumn{3}{c}{Diagnosis Decision} \\
        \cmidrule(lr){2-4}  
         & Prec. (\%) & Rec. (\%) & Acc. (\%) \\  
        \midrule
        Gemini         & \textbf{84.48} & \textbf{100.00} & \textbf{75.38} \\
        QWen                      & 47.91 & 58.33 & 53.33 \\
        GPT-4.1         & 43.48 & 77.77 & 61.54 \\
        Llama                       & 56.25 & 61.50 & 50.84 \\  
        Claude              & 53.53 & \textbf{100.00} & 65.21 \\  
        InternVL        & 46.23 & 61.50 & 53.33 \\  
        \bottomrule
    \end{tabular}
\end{table}

\subsubsection{Statistic Analysis}
To quantify the uncertainty of our performance metrics, we employed a bootstrap resampling method. The error bars presented for each model's performance in \textbf{Figure \ref{fig:asset3}} and \textbf{Figure \ref{fig:asset4}} represent the 95\% confidence intervals, which were derived from the 2.5th and 97.5th percentiles of 1,000 bootstrap iterations on the respective test sets.

To formally compare our Pathologist-o3 agent against the next-best performing models, we conducted a paired Student's t-test on the distributions of performance differences obtained from the bootstrap procedure. Across both the internal validation set and the external LNCO2 cohort, Pathologist-o3 demonstrated a statistically significant improvement in overall performance over the second-best model, with p-values consistently below the significance threshold of 0.05 (p<0.05) for key metrics including accuracy, precision, and recall.

\begin{table*}[ht]
\centering
\caption{Performance metrics of various models with 95\% confidence intervals. Best performed method(s) were highlighted in bold font.}
\label{tab:fig2_model_performance}
\begin{tabular}{lccc}
\hline
Model & Prec. (\%) & Rec. (\%) & Acc. (\%) \\
\toprule
Gemini & 33.3 (22.2 - 45.5) & 18.2 (9.1 - 30.0) & 60.0 (51.8 - 67.9) \\
o4-mini & 55.0 (42.9 - 66.7) & 45.8 (32.1 - 60.0) & 65.6 (57.1 - 73.6) \\
GPT-4.1 & 42.3 (38.8 - 45.7) & 91.7 (81.2 - 97.9) & 50.0 (41.6 - 58.4) \\
QwenVL & 37.3 (28.8 - 46.4) & 52.1 (38.3 - 65.6) & 49.2 (40.9 - 57.7) \\
Phi-4 & 41.0 (28.6 - 54.5) & 33.3 (20.8 - 47.8) & 57.0 (48.5 - 65.2) \\
Llama-4 & 75.0 (33.3 - 100.0) & 6.2 (2.1 - 14.9) & 64.1 (55.8 - 72.0) \\
InternVL & 37.7 (34.5 - 41.0) & 83.3 (70.8 - 93.8) & 42.2 (34.0 - 50.4) \\
o3 & 46.7 (42.9 - 50.0) & 87.5 (75.0 - 95.8) & 57.8 (49.3 - 66.2) \\
Ours & \textbf{84.5 (78.9 - 89.5)} & \textbf{100.0}  & \textbf{75.4 (67.5 - 82.5)} \\
\bottomrule
\end{tabular}
\end{table*}

\begin{table*}[ht]
\centering
\caption{Performance metrics on the external LNCO2 cohort with 95\% confidence intervals. Best performed method(s) were highlighted in bold font.}
\label{tab:external_validation_performance}
\begin{tabular}{lccc}
\toprule
Model & Acc. (\%) & Prec. (\%) & Rec. (\%) \\
\midrule
Gemini & 62.0 (56.7 - 67.3) & 23.5 (7.7 - 46.2) & 7.8 (2.4 - 17.1) \\
Qwen & 44.5 (39.3 - 50.2) & 10.9 (5.3 - 18.4) & 46.3 (31.7 - 61.0) \\
GPT-4o & 40.3 (34.9 - 45.8) & 9.8 (4.8 - 17.1) & 51.5 (36.6 - 65.9) \\
o3 & 40.2 (34.9 - 45.5) & 15.5 (9.3 - 23.3) & 82.9 (70.7 - 92.7) \\
Pathologist-o3 & \textbf{69.4 (64.2 - 74.5)} & \textbf{62.9 (48.6 - 77.1)} & \textbf{97.6 (87.8 - 100.0)} \\
\bottomrule
\end{tabular}
\end{table*}

\subsection{The Pathology-o3 Agent: Behavior Predictor}

To validate our dataset, we created an agent named Pathology-o3. Its core navigational component is a task-specific Behavior Predictor, which is trained to propose diagnostically relevant ROIs on a new WSI. The Behavior Predictor is not a single, monolithic model. Instead, our framework envisions a library of predictors, each trained for a distinct, high-volume clinical task (e.g., melanoma screening, prostate cancer grading). For this paper, we focus on developing and evaluating one such predictor as a proof-of-concept: a model trained for the common and time-consuming task of identifying lymph node metastasis in colorectal cancer cases.

Our choice of a YOLO-based architecture for these individual Behavior Predictors is deliberate and grounded in clinical pragmatism. For this study, the predictor is a lightweight YOLOv8~\cite{yolov5} object detection model, chosen for its efficiency and ease of deployment. Its lightweight nature ensures that the model can be readily deployed in typical hospital IT environments, including on standard CPUs or within web-based viewers. This accessibility paves the way for future enhancements like on-the-fly incremental learning, where a predictor could be continuously personalized to a pathologist's specific style.

We considered more advanced, open-vocabulary detection models (e.g., based on VLM-grounding or CLIP), but found them less suitable for this clinical application. Their "one-model-for-all" design does not align with the sub-specialized reality of pathology. The clinical need is not for a single, universal model, but for highly efficient and reliable models for specific, repetitive tasks. Our modular approach of training lightweight, task-specific predictors is more practical and robust to variations in scanning hardware and magnification—a common challenge for end-to-end models.

The Behavior Predictor was trained on the 1,521 ROI boxes extracted by the AI Session Recorder from 137 WSIs across all eight pathologists. It learns to identify areas with high diagnostic relevance based on expert behavior, which are not necessarily ``abnormal'' in a clinical sense. To ensure clinical validity and prevent data leakage, training was performed using a 5-fold cross-validation strategy, with splits made at the patient level.

For implementation, the model was trained for 100 epochs using the Adam optimizer with an initial learning rate of 1e-4 and a batch size of 32. All input images were resized to 1024x1024 pixels. To improve robustness, we applied standard data augmentation techniques, including random flips, rotations, scaling, and color adjustments. The training was guided by a standard composite loss function:



\begin{itemize}
\item \textbf{Localization Loss:} A regression loss (CIoU loss) that penalizes the discrepancy between the predicted and ground-truth bounding boxes. The Complete Intersection over Union (CIoU) loss is defined as:
$$L_{CIoU} = 1 - IoU + \frac{\rho^2(b, b^{gt})}{c^2} + \alpha v$$
where $IoU$ is the Intersection over Union, $\rho^2(b, b^{gt})$ is the squared Euclidean distance between the center points of the predicted box ($b$) and the ground-truth box ($b^{gt}$), $c$ is the diagonal length of the smallest box enclosing both boxes, and the term $\alpha v$ is a penalty term for the consistency of the aspect ratio.

\item \textbf{Classification Loss:} A classification loss (Binary Cross-Entropy) that trains the model to distinguish relevant regions from the background. The Binary Cross-Entropy (BCE) loss is calculated as:
$$L_{BCE} = -\frac{1}{N} \sum_{i=1}^{N} [y_i \log(\hat{y}_i) + (1-y_i) \log(1-\hat{y}_i)]$$
where $N$ is the number of predictions, $y_i$ is the ground-truth label (e.g., 1 for a relevant region, 0 for background), and $\hat{y}_i$ is the model's predicted probability that the region is relevant.
\end{itemize}

In the validation of the Pathology-o3 in \textbf{Figure \ref{fig:asset3}}, \textbf{Figure \ref{fig:asset8}}, \textbf{Figure \ref{fig:asset9}}, prediction from 5 test fold is merged and used as Behavior Predictor's output. 
In external validation on LNCO2 dataset (\textbf{Figure \ref{fig:asset4}}), fold-0 model is used to generate behavior bounding boxes prediction.

To evaluate how effectively our model replicates the navigation behavior of expert pathologists, we define two primary metrics: Behavior Efficiency and Behavior Completeness. These metrics assess the model's ability to predict the pathologist's next area of focus, represented by bounding boxes corresponding to viewed regions. A predicted bounding box is considered a successful identification or ``hit'' if either its Intersection over Union with an expert-visited region surpasses a predefined threshold (0.3), or if one bounding box is contained within the other. This inclusive criterion is adopted to reasonably account for differing magnification preferences among pathologists, where a region of interest might be correctly identified but captured with a bounding box of a different scale, which could otherwise lead to a low IoU despite accurate localization.

\textbf{Behavior Efficiency.} This metric quantifies the precision of the model's predicted areas of focus. It reflects how many of the model-proposed regions are indeed relevant to the expert's examination path. High Behavior Efficiency indicates that the model minimizes suggestions for irrelevant areas, thus potentially leading to a more streamlined and efficient diagnostic workflow. It is calculated as:

\[
\text{Behavior Efficiency} = \frac{\text{Number of Hits}}{\text{Total Number of Model-Predicted Regions}}
\]

\textbf{Behavior Completeness.} This metric measures the recall of the model's predictions. It assesses the proportion of expert-visited regions that the model successfully identifies. High Behavior Completeness suggests that the model comprehensively covers the areas deemed important by the expert pathologist, minimizing missed diagnostically relevant regions. It is calculated as:
\[
\text{Behavior Completeness} = \frac{\text{Number of Hits}}{\text{Total Number of Expert-Visited Regions in Ground Truth}}
\]

\subsection{Experimental Setting Details}

\subsubsection{Classification metrics.}  
For discrete prediction tasks, we report a suite of complementary classification metrics. Let $TP$ denote true positives (correctly identified positive cases), $TN$ true negatives (correctly identified negative cases), $FP$ false positives (negative cases incorrectly predicted as positive), and $FN$ false negatives (positive cases incorrectly predicted as negative).  
  
Accuracy is the most widely used classification metric, defined as the proportion of correctly classified samples among all cases:
\[
\text{Accuracy} = \frac{TP + TN}{TP + TN + FP + FN}.
\]
It provides an intuitive single-number summary of overall correctness and is easy to interpret across different tasks. However, accuracy alone can be misleading in settings with severe class imbalance, as a model may achieve deceptively high scores by always predicting the majority class. For instance, in cancer screening with 95\% healthy cases, a trivial model labeling all patients as healthy would still reach 95\% accuracy while completely failing to detect diseased individuals. These limitations motivate the complementary use of precision, recall, and F1-score, which better capture diagnostic reliability under clinical imbalance.
 
To complement accuracy, we report precision, recall, and their harmonic mean (F1-score), which together provide a more nuanced assessment of diagnostic performance under class imbalance. 

Precision quantifies the reliability of positive predictions, indicating the proportion of predicted positives that are truly positive.
\[
\text{Precision} = \frac{TP}{TP + FP}.
\]
A high precision indicates that most predicted positives are true positives, thereby reducing the risk of over-diagnosis and unnecessary clinical interventions.  

Recall (sensitivity) captures the model’s ability to identify all true positive cases, thereby minimizing false negatives. This is crucial in clinical contexts such as cancer detection, where missed diagnoses can delay treatment. 
\[
\text{Recall} = \frac{TP}{TP + FN}.
\]
High recall is essential in clinical contexts such as early cancer detection, where missing positive cases (false negatives) could delay treatment and lead to adverse outcomes.  

\subsubsection{Comparison to State-of-the-Art Models}
To benchmark our agent, we compared its diagnostic performance on the CRC lymph node task against a panel of leading VLMs (\textbf{Figure~\ref{fig:asset3}c}). These baseline models were evaluated on their ability to perform the same task without the guidance of our Behavior Predictor. For reference we also provide their general vision capability in \textbf{Table \ref{tab:mmmu_ocr_overlap}}. Better performance on general vision task not results in more accurate diagnosis (Llama 4 vs Qwen). Full specifications of the models used and the dates of API access are detailed in \textbf{Table \ref{tab:vlm_specifications}}. 

\subsubsection{External Validation}
To assess generalization, we performed a rigorous out-of-distribution test using the independent LNCO2 cohort (\textbf{Figure~\ref{fig:asset4}}). This dataset from different continent (Sweden) contains images from different scanners (\textit{Leica Aperio} and \textit{Hamamatsu}) and at different magnifications (20x/40x), presenting a significant domain shift. (All slides from training cohort were scanned by Leica Aperio Scanner with 40x magnification at 0.25 µm per pixel). The Behavior Predictor was trained exclusively on our single-center Stanford data and then evaluated end-to-end on the LNCO2 dataset without any fine-tuning.

\subsubsection{Behavioral Analyses}
To deconstruct agent performance (\textbf{Figure~\ref{fig:asset8}}), we conducted several analyses using the ground truth navigation path of a senior attending pathologist.
\begin{itemize}
    \item \textbf{Robustness to Viewing Order:} The complete set of expert-selected ROIs for a case was presented to the agent in its original sequence (Forward), in reverse (Reverse), and in a randomly shuffled order (Random).
    \item \textbf{Effect of Evidence Quantity:} To simulate a Chain-of-Thought of varying length, we capped the maximum number of ROIs the agent could inspect at a given limit.
    \item \textbf{Unguided Navigation:} We evaluated the ability of unguided VLMs to navigate by tasking them to select their own ROIs. We measured their performance using two metrics: \textbf{completeness} (recall), defined as the fraction of expert-viewed ROIs they successfully identified, and \textbf{efficiency} (precision), the fraction of their proposed ROIs that were relevant. A prediction was considered a successful hit if its IoU with an expert ROI was > 0.3.
\end{itemize}

\subsubsection{Practicality and Scalability Analysis}
To assess practical viability (\textbf{Figure \ref{fig:asset9}}), we measured the per-slide cost and latency of our agent using two different VLMs: a high-performance model (Gemini 2.5 Pro) and a more cost-effective model (Gemini 2.5 Flash). Our cost calculations are based on public API pricing at the time of experimentation, which typically charges for both input and output tokens, with output tokens being more expensive. For image inputs, we standardized all ROIs to a 1024x1024 resolution to ensure consistent pricing per region, as many models charge based on pixel count. This analysis demonstrates that by carefully managing prompt length and standardizing image inputs, an agentic workflow can be made cost-effective, particularly when using smaller, faster models for less complex tasks. To analyze scalability, we compared three conditions across multiple VLM backbones: the baseline VLM without guidance, our agent guided by its \emph{learned} behavior, and an oracle agent guided by the \emph{real} behavior of an expert pathologist.

\section*{Author Contributions}
S.W. and Z.H. conceived the study. S.W., R.W and Y. L. developed the methodology. S.W. and R.W. performed the experiments. S.W., J.S., and Z.H. curated and acquired the data. C.H., S.K., and J.S. provided expert pathology annotation and feedback. S.W. and Z.H. wrote the manuscript. Z.H. supervised the study. All authors discussed the results and approved the final manuscript.

\section*{Data Availability}

Processed behavior data from 8 pathologists on the Stanford CRC lymph node dataset, including all intermediate bounding boxes, cropped ROIs, slide thumbnails, and the Chain-of-Thought dataset, will be made publicly available at \href{https://github.com/zhihuanglab/Pathology-CoT}{https://github.com/zhihuanglab/Pathology-CoT}.
Whole-slide images and expert annotations from the LNCO2 dataset are available via the AIDA Data Hub at \url{https://datahub.aida.scilifelab.se/10.23698/aida/lnco2} upon application and approval.

\section*{Code Availability}
Source code is publicly available at \href{https://github.com/zhihuanglab/Pathology-CoT}{https://github.com/zhihuanglab/Pathology-CoT}.

\section*{Conflict of Interests}
None declared.

\newpage
\bibliography{ref.bib}

\newpage
\appendix

\section{Appendix}
\setcounter{figure}{0}
\renewcommand{\thefigure}{S\arabic{figure}}
\begin{table}[t]
\centering
\small
\begin{tabular}{lrrrr}
\toprule
\textbf{Model} & \textbf{MMMU (Val) $\uparrow$} & \textbf{Prec. (\%) $\uparrow$} & \textbf{Rec. (\%) $\uparrow$} & \textbf{Acc. (\%) $\uparrow$} \\
\midrule
o3                       & \textbf{82.9} & 46.7 & \textbf{87.5} & 57.8 \\
o4-mini                  & 81.6 & 55.0 & 45.8 & \textbf{65.6} \\
Llama 4 Maverick         & 73.4 & \textbf{75.0} & 6.2  & 42.2 \\
Gemini 2.5 Pro 05-06     & 79.6 & 33.3 & 18.2 & 60.0 \\
Qwen-VL-PLUS             & 45.2 & 37.3 & 52.1 & 49.2 \\
\bottomrule
\end{tabular}
\caption{Overlap between VLM general capbility (MMMU) overall scores and diagnosis Precision/Recall/Accuracy. General vision language capability have domain gap to pathology image diagnosis. Better performance on general vision task not results in more accurate diagnosis. MMMU results are from \href{https://mmmu-benchmark.github.io}{https://mmmu-benchmark.github.io}}
\label{tab:mmmu_ocr_overlap}
\end{table}

\subsection{The AI Session Recorder Behavior Analyze Pipeline}
\label{sec:supp_behavior_analyze}

\begin{figure}[hbtp]
    \centering
    \includegraphics[width=1\linewidth]{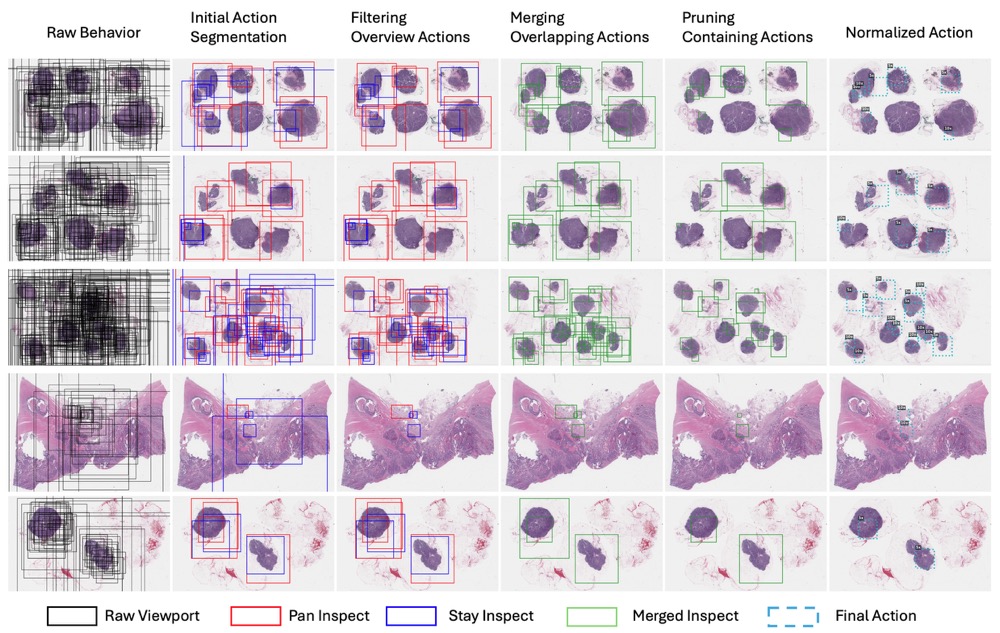}
    \caption{\textbf{The Behavior Analyze Pipeline for Preprocessing Pathologist Navigation.} The process transforms raw, high-frequency viewport data into a clean set of actions. From left to right: \textbf{(1) Raw Behavior:} The initial noisy data with many overlapping viewports. \textbf{(2) Initial Action Segmentation:} Raw events are grouped into \texttt{StayInspect} (blue) and \texttt{PanInspect} (red) actions based on temporal heuristics. \textbf{(3) Filtering Overview Actions:} Very large, low-magnification bounding boxes corresponding to non-diagnostic overviews are removed. \textbf{(4) Merging Overlapping Actions:} Spatially-proximal actions with high IoU are merged into single, consolidated actions. \textbf{(5) Pruning Containing Actions:} Larger, redundant actions that fully contain smaller, more specific inspection actions are pruned. \textbf{(6) Normalizing Action Bounding Boxes:} The final set of actions is resized to standard dimensions corresponding to discrete magnification levels (e.g., 5x, 10x), creating a consistent input format for the VLM.}
    \label{fig:pipeline_detail}
\end{figure}

In this section, we provide a detailed, step-by-step technical description of the \textbf{Behavior Analyze} pipeline used to preprocess raw pathologist interaction logs into a structured, VLM-friendly format. The entire pipeline is visually summarized in \textbf{Figure~\ref{fig:pipeline_detail}}, which illustrates the transformation of noisy, overlapping raw viewport data into a clean, normalized set of action bounding boxes. The implementation details described below correspond to the logic in the provided code.

\begin{figure}[hbtp]
    \centering
    \includegraphics[width=1\linewidth]{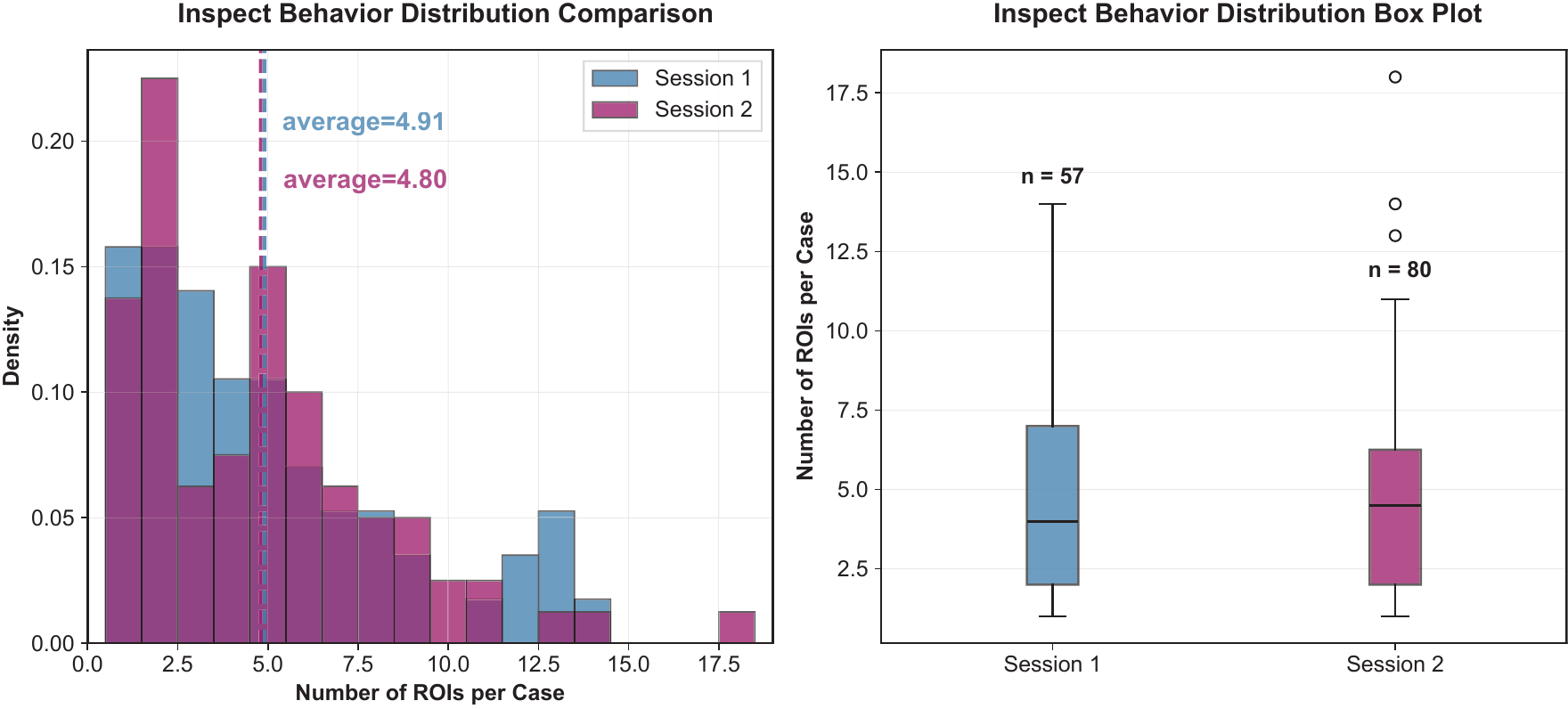}
    \caption{Comparison of AI Session Recorder generated \textbf{Inspect} ROI results from two sessions of same attending pathologist. Same pathologist generate very similar results across different session, showing the stability and reliability of AI Session Recorder. }
    \label{fig:session_compare}
\end{figure}

\begin{figure}[hbtp]
    \centering
    \includegraphics[width=1\linewidth]{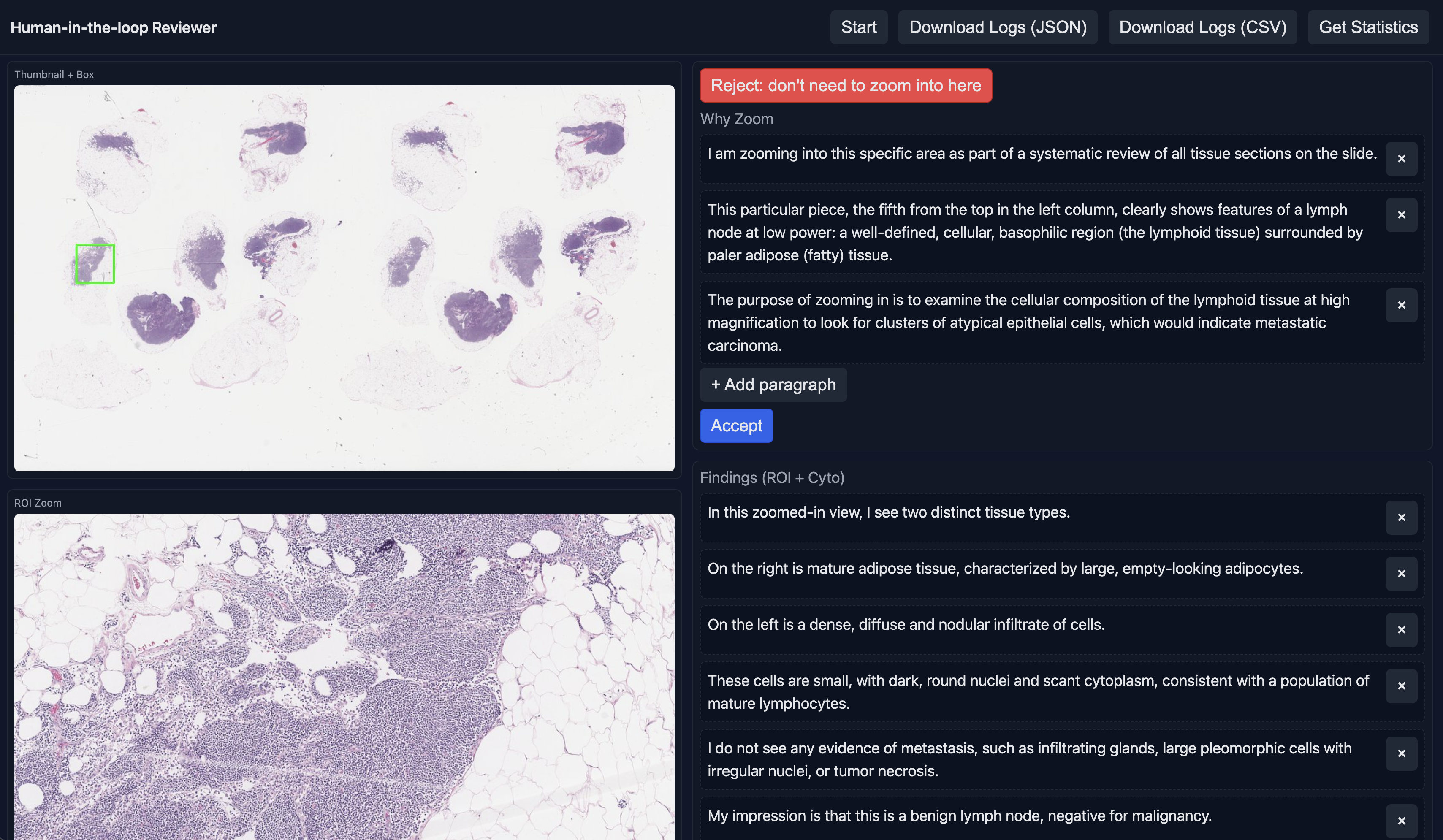}
    \caption{A software for pathology ROI review presents the slide thumbnail with the selected ROI box, a zoomed ROI view, and an optional cytology crop. AI‑drafted text for “Thumbnail Impression,” “Why Zoom,” and “Findings (ROI + Cyto)” is shown for the pathologist to edit and either Accept or Reject; pressing “Next” advances and, if not rejected, treats the ROI as accepted. The header displays case ID, ROI index, progress, elapsed time, and decision state. }
    \label{fig:human_in_loop_review_software}
\end{figure}

\begin{figure}[hbtp] 
    \centering
    \includegraphics[width=1\linewidth]{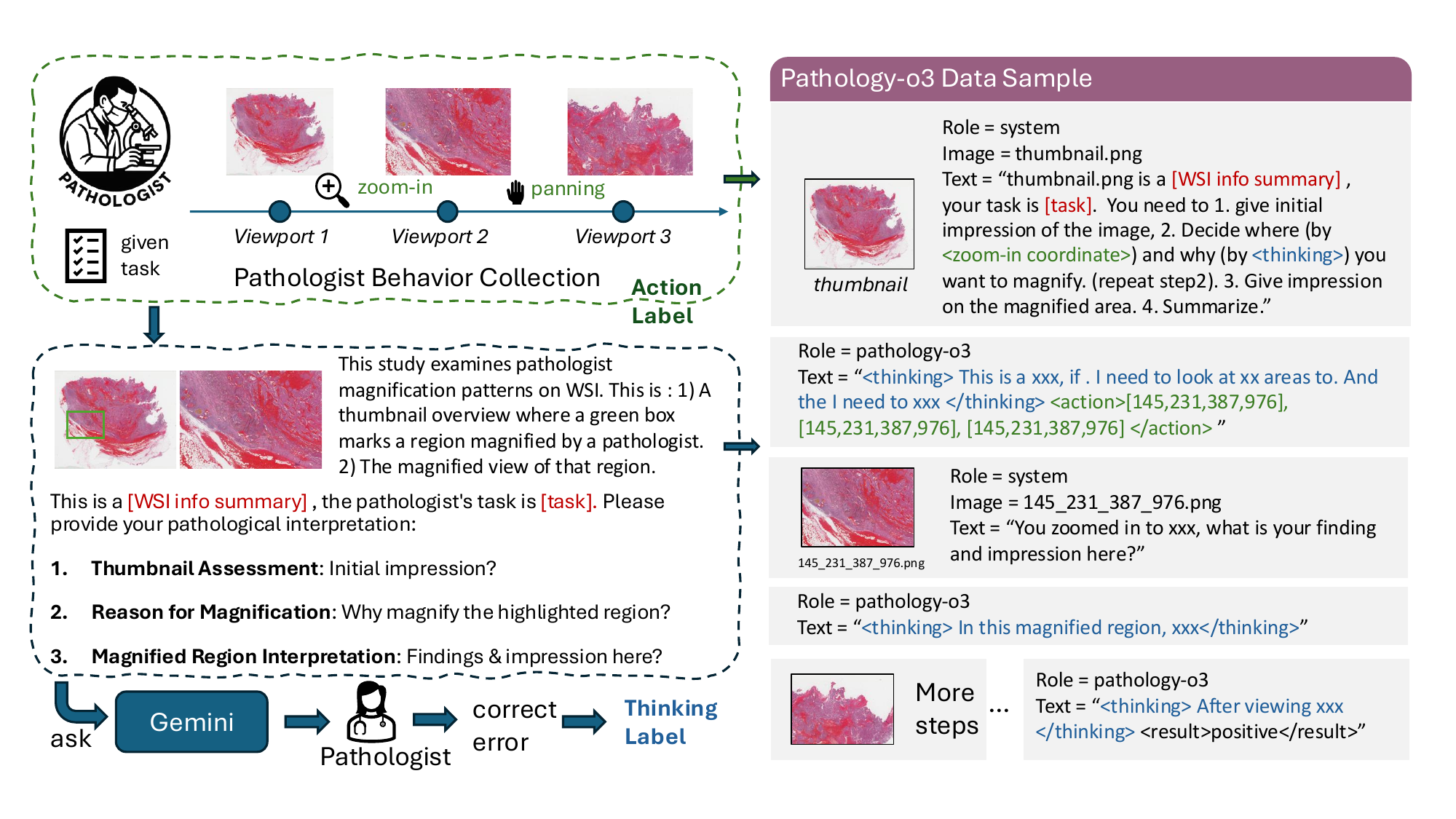} 
    \caption{\textbf{Pathology-CoT Dataset Creation Pipeline.} Left: Pathologist interaction data (viewports, navigation actions like zoom-in/panning, task context) is collected. Center: Generating the ``Thinking Label''. Contextual information (task, current view, next action) is used to prompt a large language model (LLM, e.g., Gemini) to generate candidate reasoning. A pathologist then reviews, corrects, or approves this output to create the final ``Thinking Label''. Right: The collected actions and generated thinking labels are merged with corresponding image patches (thumbnails, zoomed regions) and system prompts into a structured conversational format suitable for model training or in-context learning.}
    \label{fig:dataset_creation} 
\end{figure}

\begin{figure}[hbtp]
    \centering
    \includegraphics[width=\linewidth]{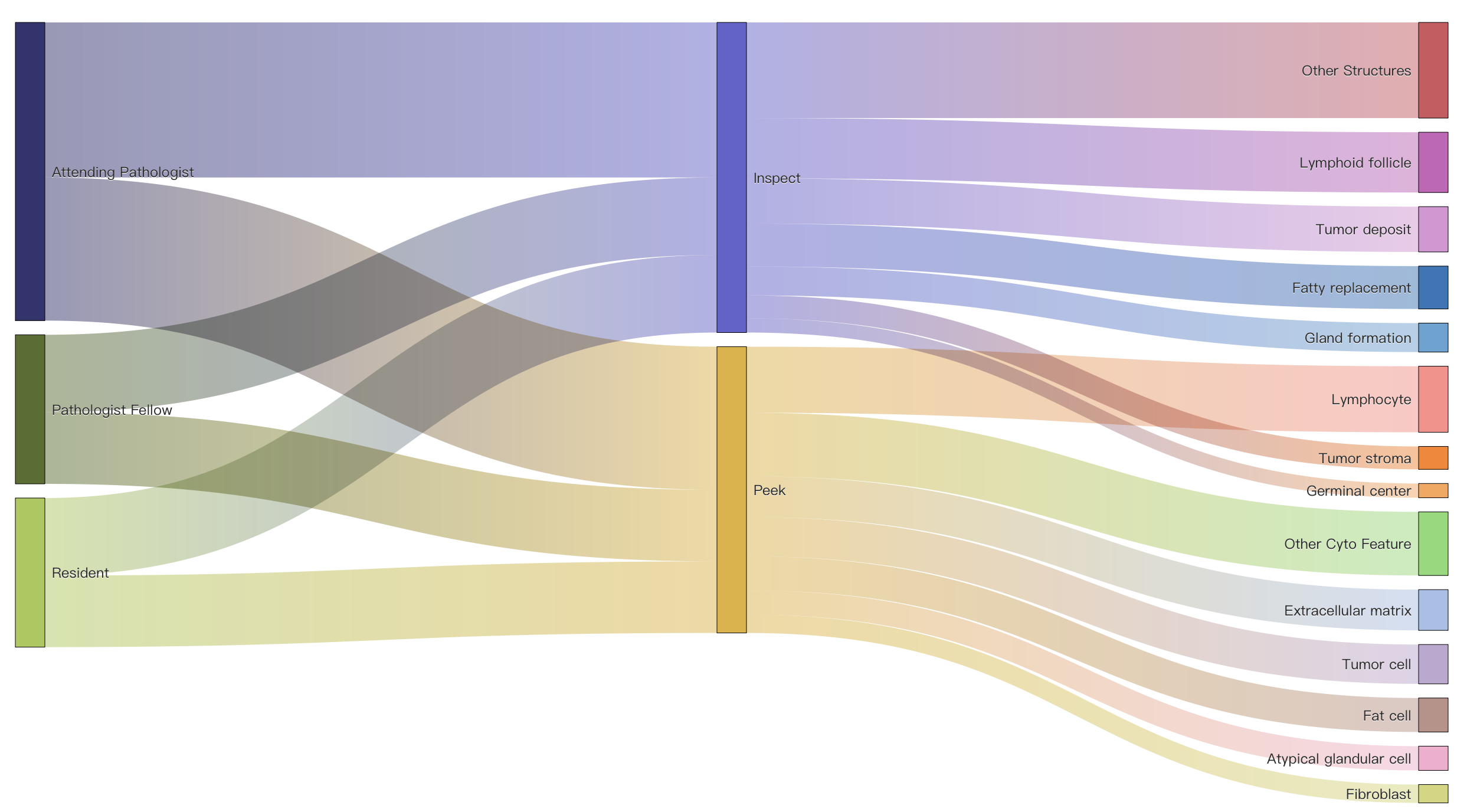}
    \caption{\textbf{Sankey Diagram of Pathology-CoT Data Distribution} The diagram illustrates how two ROIs, Inspect (broad-field scan) and Peek (high-magnification look), are applied to analyze various histological and cytological features during lymph node metastasis diagnosis. The flow size represents the relative proportion of diagnostic attention directed from a specific method to each feature.}
    \label{fig:sankey_analyze}
\end{figure}

\subsubsection{Timing study design and fairness adjustment}

We quantified the efficiency of the AI Session Recorder + human-in-the-loop review relative to creating labels from scratch using a controlled timing study. Two users (User~1 and User~2) reviewed AI Session Recorder-drafted text for each region of interest (ROI) using the interface in \textbf{Figure~\ref{fig:human_in_loop_review_software}}. For every ROI, the system recorded wall-clock time from ROI load to decision. We summarize three per-round quantities: \emph{Verify} (accepted with no edits), \emph{Revise} (accepted after any edit), and their across-round \emph{Average}. The experiment comprised 10 representative cases and 78 total conversational rounds.

To align with the chain-of-thought supervision used in the main task, users in the manual baselines were instructed to provide both a final diagnosis (e.g., lymph-node positive vs. negative) and a short rationale describing (i) abnormal/atypical features supporting the decision and (ii) normal structures present. Most AI Session Recorder drafts were accepted without edits; the observed revision rates were 15.59\% for User~1 and 21.65\% for User~2, consistent with over 80\% of AI-generated text requiring no edits.

Exact reproduction of the source pathologist's navigation is infeasible for new users, which would confound per-round timing if reviewers were allowed to free-roam. To ensure comparability, we fixed the items of interest to the source pathologist's ROIs for all conditions (verify/revise and manual baselines). Because this anchoring removes the navigation effort that typically precedes manual labeling, manual baselines were adjusted to include the source expert's recorded navigation time when locating each ROI. For round \(r\) we compute
\[ T_{\mathrm{manual}}^{(r)} = T_{\mathrm{write}}^{(r)} + T_{\mathrm{nav}}^{\mathrm{expert},(r)}, \]
where \(T_{\mathrm{write}}\) is the time a user spent composing text from scratch (typing or dictation) for the fixed ROI and \(T_{\mathrm{nav}}^{\mathrm{expert}}\) is the original pathologist's navigation time to that ROI. This yields a conservative, apples-to-apples comparison with the verify/edit workflow, which requires no new navigation.

Finally, we note a consistent verbosity gradient across conditions: free text written from scratch was shorter when typing than when dictating, and both were shorter than the AI drafts presented for verification. The latter are intentionally comprehensive to minimize missed details during review.

\subsubsection{Reviewer interface design}
\label{sec:reviewer_ui_design}
The reviewer tool (\textbf{Figure.~\ref{fig:human_in_loop_review_software}}) was engineered to minimize edit latency while preserving an editable record of decisions. The layout presents, from left to right, a slide thumbnail with the selected ROI box, a zoomed ROI view (and an optional cytology crop), and editable text panels for ``Thumbnail Impression,'' ``Why Zoom,'' and ``Findings (ROI + Cyto).'' The header shows case identifier, ROI index, progress, elapsed time, and the current decision state.

AI-drafted rationales are segmented at sentence boundaries on load, and each sentence is rendered as an individual paragraph with an adjacent one-click delete control (\(\times\)). Typical drafts contain five to seven sentences. This sentence-level granularity enables ``surgical'' edits—reviewers can remove a single over-claiming or inaccurate sentence without retyping the remainder—thereby reducing total revision time and contributing to the low observed revision rates. Global controls allow the reviewer to \emph{Accept} the edited draft or \emph{Reject} the ROI when zoom-in is deemed unnecessary; advancing to the next ROI records the decision and timing metadata automatically.

\subsubsection{Step 1: Initial Action Segmentation}
\textbf{Goal:} To convert the high-frequency stream of viewport events into an initial set of meaningful, low-level actions.

\textbf{Method:} This step is handled by the \texttt{BehaviorProcessor} class in our script. It iterates through the sequence of timestamped viewport events (\texttt{record.zoom\_times}, \texttt{record.zoom\_chain}). It uses predefined temporal thresholds to identify two foundational action types:
\begin{itemize}
    \item \textbf{Stay Inspect:} A static inspection is identified if the dwell time in a single viewport exceeds a threshold (set to 1.0s).
    \item \textbf{Pan Inspect:} A panning inspection is identified if a sequence of 'pan' events occurs continuously for a cumulative duration exceeding a threshold (set to 2.0s).
\end{itemize}
For each identified action, a bounding box is created. For a \texttt{StayInspect}, the box is centered on the viewport. For a \texttt{PanInspect}, a minimal bounding box is calculated to encompass all viewports within the continuous panning sequence. The result is an initial list of \texttt{VLMAction} objects, as shown in the ``Stay \& Pan'' column of \textbf{Figure~\ref{fig:pipeline_detail}}.

\subsubsection{Step 2: Filtering Overview Actions}
\textbf{Goal:} To remove non-diagnostic, low-magnification overview actions that do not represent detailed inspection.

\textbf{Method:} This step is performed by the \texttt{filter\_big\_bboxes()} function. Pathologists often zoom out to get their bearings, resulting in extremely large bounding boxes covering a substantial portion of the WSI. These actions are not useful for training a model to find specific ROIs. The function filters the action list by removing any \texttt{VLMAction} whose bounding box width is greater than a set fraction (e.g., 40\% or 2/5) of the total WSI height (\texttt{wsi\_height}). This effectively removes the ``birds-eye view'' actions, focusing the dataset on genuine inspection events. This is depicted in the ``Remove Big'' column of \textbf{Figure~\ref{fig:pipeline_detail}}.

\subsubsection{Step 3: Merging Overlapping Actions}
\textbf{Goal:} To consolidate multiple, highly-overlapping actions that likely correspond to a single cognitive task of inspecting one larger region.

\textbf{Method:} The \texttt{merge\_similar()} function implements this logic. It iteratively compares all pairs of actions in the current list. If the Intersection over Union (IoU) between two action bounding boxes (\texttt{calculate\_iou()}) exceeds a defined \texttt{iou\_threshold} (e.g., 0.5-0.8), the two actions are considered part of the same inspection event. They are then merged into a single new \texttt{VLMAction} (\texttt{merge\_two\_actions()}). The new action's bounding box is the union of the two original boxes, and its duration spans the full time from the start of the first action to the end of the second. This process repeats until no more merges are possible, resulting in a smaller set of more distinct actions, as seen in the ``Merge'' column of \textbf{Figure~\ref{fig:pipeline_detail}}.

\subsubsection{Step 4: Pruning Redundant Containing Actions}
\textbf{Goal:} To refine the action list by removing larger, less specific actions that are made redundant by smaller, more detailed inspections occurring within them.

\textbf{Method:} This refinement is handled by the \texttt{filter\_mostly\_contained\_actions()} function. It addresses cases where a pathologist inspects a large area (e.g., a lymph node) and then zooms in to inspect a specific feature within it (e.g., a cluster of tumor cells). In this scenario, the smaller, high-magnification action is more informative. The function compares every pair of actions. If a smaller action's bounding box is mostly contained within a larger action's box (i.e., the intersection area is >90\% of the smaller box's area, checked by \texttt{check\_containment\_by\_area()}), the larger, containing action is marked as redundant and removed from the list. This prioritizes the most detailed and specific points of interest identified by the pathologist, as illustrated in the ``Find Include'' column of \textbf{Figure~\ref{fig:pipeline_detail}}.

\subsubsection{Step 5: Normalizing Action Bounding Boxes}
\textbf{Goal:} To standardize the final action bounding boxes into a consistent format, mimicking discrete microscope objectives.

\textbf{Method:} The final step, \texttt{standardize\_action\_bboxes()}, ensures that the input to the VLM is uniform. It maps each action's bounding box to one of two standard sizes based on its original area, corresponding to a ``5x'' (larger box) or ``10x'' (smaller box) view. The function calculates the center of the final, pruned bounding box and resizes it to the standard dimension (e.g., \texttt{wsi\_height / 5.0} for a 5x view). This process also re-bins the action's magnification label (\texttt{magnification\_bin}) to match the standardized size. The output is a clean, normalized set of action bounding boxes of consistent sizes, centered on the pathologist's key areas of focus. This final, structured output is shown in the ``Normalize'' column of \textbf{Figure~\ref{fig:pipeline_detail}} and is ready for use in generating ground truth for model training.

\subsection{Prompts for VLM Analysis}
This section details the specific prompts used to guide the Vision-Language Model (VLM) at different stages of the diagnostic workflow.

\begin{tcolorbox}[title=Prompt for Initial WSI Overview Analysis,colback=blue!10!white,colframe=violet!60!black,fonttitle=\bfseries]
This is an H\&E WSI of a CRC case. The task is to find all positive lymph nodes.
An expert pathologist has identified the regions of interest marked in the image for closer examination.
What is your initial impression of the overall image?
Format your response as: \texttt{<impression>your overall impression of the slide</impression>}
\end{tcolorbox}

\begin{tcolorbox}[title=Prompt for Region of Interest Analysis ,colback=green!5!white,colframe=green!60!black,fonttitle=\bfseries]
What is your impression on this region?
Second image is a maximum-magnification crop from the center of the region, intended for observing fine cytological features such as nuclei and cytoplasm.
Please consider that space between cells is not always cytoplasm, but could be an artifact from processing.
Lymph node that is dead or completely occupied by tumor / or dead tumor cell, this should be called tumor deposit not a positive lymph node.
\end{tcolorbox}

\begin{tcolorbox}[title=Prompt for Final Summary and Diagnosis Generation ,colback=red!5!white,colframe=red!70!black,fonttitle=\bfseries]
Please provide a comprehensive final pathological impression and diagnosis based on all the above analyses. Consider:
\begin{enumerate}
    \item The overall tissue architecture and morphology
    \item Findings from each specific region
    \item Any patterns or correlations between regions
    \item Your final diagnostic impression, please do not consider suspicious region as positive
\end{enumerate}

Format your response EXACTLY as follows:

\texttt{<final\_impression>}Your comprehensive final diagnostic impression\texttt{</final\_impression>}

\texttt{<recommendations>}Any additional recommendations\texttt{</recommendations>}

\texttt{<diagnostic\_info>}

PT\_or\_LN: ``PT'' if this is a primary tumor section, ``LN'' if this is a lymph node section

t\_stage: [1-4] if primary tumor, 0 if lymph node

lymph\_node\_positive: true/false

positive\_regions: if lymph\_node\_positive is true, give a [1,2,3] like list. if false just say []

suspicious\_regions: []

\texttt{</diagnostic\_info>}
\end{tcolorbox}

\subsection{Method for Structuring Semi-Automated Thinking Labels}
\label{sec:supp_thinking_structuring}

To transform the semi-automated, free-text `Thinking Labels` into structured, analyzable data, we developed a classification pipeline powered by a large language model (\texttt{models/gemini-2.5-flash}). This method processes each thinking label to assign multiple, relevant pathological tags for both a low-magnification (regional) view and a high-magnification (cellular) view simultaneously.

The core of this method is a carefully designed prompt that instructs the LLM to perform two multi-label classification tasks in a single API call. The model is required to return the results as two comma-separated lists, delimited by a pipe character (`|`). This structured output format is crucial for robust parsing and minimizes the need for complex post-processing.

The classification taxonomy was designed to be comprehensive for colorectal cancer lymph node metastasis. For the \textbf{low-magnification (box-level)} view, classes are grouped into three categories: (1) \textit{Tumor-related features} (e.g., ``Gland formation'', ``Necrosis''), (2) \textit{Normal Lymph Node Structures} (e.g., ``Germinal center'', ``Sinus''), and (3) \textit{Reactive or Pathological Changes} (e.g., ``Fibrosis'', ``Hemorrhage''). This allows for capturing the overall tissue architecture. For the \textbf{high-magnification (40x-level)} view, classes are also grouped into three categories: (1) \textit{Tumor Cell-related features} (e.g., ``Tumor cell'', ``Mitotic figure''), (2) \textit{Normal Background Cells and Structures} (e.g., ``Lymphocyte'', ``Extracellular matrix''), and (3) \textit{Pathological Processes} (e.g., ``Apoptotic body'', ``Dead cell''). This captures fine-grained cytological details.

The specific prompt used for this pipeline is detailed below.

\begin{tcolorbox}[title=Prompt for Multi-Label Classification of Thinking Labels,colback=blue!10!white,colframe=violet!60!black,fonttitle=\bfseries]
Based on the conversation text, classify what is observed in this ROI for both low-magnification (box) and high-magnification (40x) views.

This is a MULTI-LABEL classification - multiple classes can be present simultaneously.

Box classes (low-magnification): Tumor deposit, Gland formation, Tumor stroma, Necrosis, Germinal center, Lymphoid follicle, Medullary cord, Sinus, Paracortex, Sinus histiocytosis, Fibrosis, Congestion, Hemorrhage, Fatty replacement, other
40x classes (high-magnification): Tumor cell, Mitotic figure, Atypical glandular cell, Signet ring cell, Lymphocyte, Plasma cell, Macrophage, Endothelial cell, Fibroblast, Erythrocyte, Fat cell, Extracellular matrix, Apoptotic body, Dead cell, Inflammatory cell, other

Return ONLY the two classification results in this format:

box\_classifications|40x\_classifications

Where each classification is a comma-separated list of applicable classes.

Conversation text: {text}

Instructions:
- Choose ALL applicable classes for each view from the provided lists
- Multiple classes can be present simultaneously (e.g., ``Gland formation,Fibrosis'')
- If none match well, use ``other''
- Return only the two classification lists separated by |
\end{tcolorbox}

\subsection{Specification of Evaluated Vision-Language Models}
\label{sec:supp_vlm_specs}

The Vision-Language Models (VLMs) evaluated in our experiments comprise both closed-source and open-source models. When evaluating closed-source models provided as a service through an API (e.g., from Google and OpenAI), ensuring reproducibility is a significant challenge. These models are often updated by the provider companies without public versioning, which can lead to performance shifts over time. To address this, we document the specific access timeframes for these models. For open-source or specifically versioned models, the model identifier itself is sufficient for reproducibility. The table below provides these details for each model abbreviation used in our figures.

\begin{table}[h!]
\centering
\caption{Detailed specification of the VLM backbones used for comparison experiments.}
\label{tab:vlm_specifications}
\begin{tabular}{ll}
\toprule
\textbf{Abbreviation} & \textbf{Model and Access Details} \\
\midrule
Gemini              & Official Google API, accessed in June 2025. \\
Grok                & \texttt{grok-2-vision-1212}. \\
4o-mini             & Official OpenAI API, accessed in June 2024. \\
Claude              & \texttt{claude-3.5-sonnet-20240620}. \\
o4-mini             & Official OpenAI API, accessed in July 2025. \\
GPT-4.1             & Official OpenAI API, accessed in July 2025. \\
Qwen-VL             & \texttt{Qwen-VL-Plus}. \\
Phi-4               & \texttt{microsoft/phi-4-multimodal-instruct}. \\
Llama-4             & \texttt{llama-4-maverick}. \\
InternVL            & \texttt{internvl3-14b}. \\
o3                  & Official OpenAI API, accessed in July 2024. \\
\bottomrule
\end{tabular}
\end{table}

\end{document}